\documentclass[twocolumn]{svjour3}          % twocolumn
\smartqed  % flush right qed marks, e.g. at end of proof
\usepackage{graphicx}
\usepackage{natbib}
\usepackage{times}
\usepackage{epsfig}
\usepackage{amsmath}
\usepackage{amssymb}
\usepackage{lineno}

\usepackage{xspace}
\usepackage{color}
\usepackage{multirow}
\usepackage{adjustbox}
\usepackage{subfigure}
\usepackage{paralist} % compactitem, compactenum
\usepackage{enumitem} % adjust enum margin
\usepackage{booktabs} % Publication quality tables
\usepackage{array}

\graphicspath{{figures/}}

\usepackage[pagebackref=true,breaklinks=true,letterpaper=true,colorlinks,citecolor=blue,linkcolor=blue,bookmarks=false]{hyperref}

\newcommand{\Paragraph}[1]{\vspace{-2mm}\noindent\paragraph{\normalfont\textbf{#1}}} % avoid indent for paragraph title

\long\def\ignorethis#1{}

\definecolor{orange}{rgb}{1,0.55,0}
\definecolor{gray}{rgb}{0.35,0.35,0.35}
\definecolor{MyBlue}{rgb}{0,0.2,0.8}
\definecolor{MyRed}{rgb}{0.8,0.2,0}
\definecolor{MyGreen}{rgb}{0.0,0.5,0.1}
\definecolor{MyGray}{rgb}{0.4,0.4,0.4}

\def\red#1{\textcolor{red}{#1}}
\def\blue#1{\textcolor{blue}{#1}}

\def\first#1{\red{\textbf{#1}}}
\def\second#1{\blue{\underline{#1}}}

\newlength\paramargin
\newlength\figmargin
\newlength\secmargin

\newcolumntype{L}[1]{>{\raggedright\let\newline\\\arraybackslash\hspace{0pt}}m{#1}}
\newcolumntype{C}[1]{>{\centering\let\newline\\\arraybackslash\hspace{0pt}}m{#1}}
\newcolumntype{R}[1]{>{\raggedleft\let\newline\\\arraybackslash\hspace{0pt}}m{#1}}

\newcommand{\secref}[1]{Section~\ref{sec:#1}}
\newcommand{\figref}[1]{Fig.~\ref{fig:#1}}
\newcommand{\tabref}[1]{Table~\ref{tab:#1}}
\newcommand{\eqnref}[1]{\eqref{eq:#1}}

\usepackage[pagebackref=true,breaklinks=true,letterpaper=true,colorlinks,bookmarks=false,citecolor=blue,linkcolor=blue]{hyperref}
\begin{document}

\title{Exploiting Semantics for Face Image Deblurring%\thanks{Grants or other notes
}


\author{Ziyi Shen      \and
        Wei-Sheng Lai    \and
        Tingfa Xu        \and
        Jan Kautz       \and
        Ming-Hsuan Yang    %etc.
}


\institute{Ziyi Shen
%(Corresponding author) 
\at
              Inception Institute of Artificial Intelligence, UAE and School of Optics and Photonics, Beijing Institute of Technology \\
              \email{ziyishen@bit.edu.cn}           %  \\
           \and
           Wei-Sheng Lai \and Ming-Hsuan Yang\at
              School of Engineering, University of California at Merced\\
              \email{\{wlai24,mhyang\}@ucmerced.edu}
           \and
           Tingfa Xu (Corresponding author)  \at
           	School of Optics and Photonics, Beijing Institute of Technology\\
           	\email{ciom{\_}xtf1@bit.edu.cn} 
           \and
            Jan Kautz \at Nvidia\\
            \email{jkautz@nvidia.com}
           \and
}

\date{Received: date / Accepted: date}

\maketitle

\begin{abstract}
In this paper, we propose an effective and efficient face deblurring algorithm by exploiting semantic cues via deep convolutional neural networks.
%(CNNs).
%
As the human faces are highly structured and share unified facial components (e.g., eyes and mouths), such semantic information provides a strong prior for restoration.
We incorporate face semantic labels as input priors and propose an adaptive structural loss to regularize facial local structures within an end-to-end deep convolutional neural network.
Specifically, we first use a coarse deblurring network to reduce the motion blur on the input face image.
We then adopt a parsing network to extract the semantic features from the coarse deblurred image.
Finally, the fine deblurring network utilizes the semantic information to restore a clear face image.
We train the network with perceptual and adversarial losses to generate photo-realistic results.
The proposed method restores sharp images with more accurate facial features and details.
Quantitative and qualitative evaluations demonstrate that the proposed face deblurring algorithm performs favorably against the state-of-the-art methods in terms of restoration quality, face recognition and execution speed.


\end{abstract}

\section{Introduction}
\label{intro}

Single image deblurring aims to recover a clear image from a single blurred input.
Conventional methods formulate the blur process (assuming spatially invariant blur) as the convolution operation between a latent clear image and a blur kernel, and solve this problem based on the maximum a posteriori (MAP) framework.
As the problem is ill-posed, the state-of-the-art algorithms typically rely on natural image priors (e.g., $L_0$ gradient~\citep{DBLP:conf/cvpr/XuZJ13} and dark channel prior~\citep{DBLP:conf/cvpr/PanSP016}) to constrain the solution space.

While existing image priors are effective for deblurring natural images, the underlying assumption may not hold well for images from specific categories, e.g., text, face and low-light conditions.
Numerous approaches exploit domain specific visual information, such as designing $L_0$ intensity \citep{DBLP:journals/pami/PanHS017} priors for text images or detecting light streaks \citep{Hu-CVPR-2014} for extremely low-light images.
As face images contain fewer textures and edges for estimating blur kernels, \citet{DBLP:conf/eccv/PanHSY14} search for similar face exemplars from an external dataset and extract the contour as reference edges.
However, reference images may not always exist for a specific input due to diversity of real-world face images.
Furthermore, those methods based on the MAP framework typically entail heavy computational cost due to the iterative optimization process to determine latent images and blur kernels.
The long execution time limits the applications on resource-sensitive platforms, e.g., 
mobile devices.

In this work, we propose an efficient and effective solution to deblur face images via deep convolutional neural networks (CNNs).
Since face images are highly structured and composed of similar components, 
semantic information can provide a strong prior for restoration.
We propose to leverage the face semantic labels as global priors and local constraints to train a deep CNN.
The proposed model consists of three sub-networks: a coarse deblurring network, a face parsing network, and a fine deblurring network.
The coarse deblurring network first predicts a deblurred image from the given input blurred image.
The face parsing network then estimates the semantic labels from the coarse deblurred image.
Finally, the fine deblurring network takes the blurred image, coarse deblurred image, and  semantic labels to restore a clear face image.
To encourage the network to restore fine details, we propose an adaptive local structural loss on important face components (e.g., eyes, noses, and mouths).
Finally, we impose a perceptual loss~\citep{Johnson-ECCV-2014} and an adversarial loss~\citep{GAN} to generate photo-realistic deblurred results.
As our method is end-to-end without any blur kernel estimation or post-processing, the execution time is significantly shorter than the conventional MAP-based approaches.

\begin{figure}[t]
	\footnotesize
	\renewcommand{\tabcolsep}{1pt} % adjust horizontal space
	\renewcommand{\arraystretch}{0.5} % adjust vertical space
	\begin{center}
		\begin{tabular}{ccccc}
            \includegraphics[width=0.19\columnwidth]{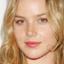} &
            \includegraphics[width=0.19\columnwidth]{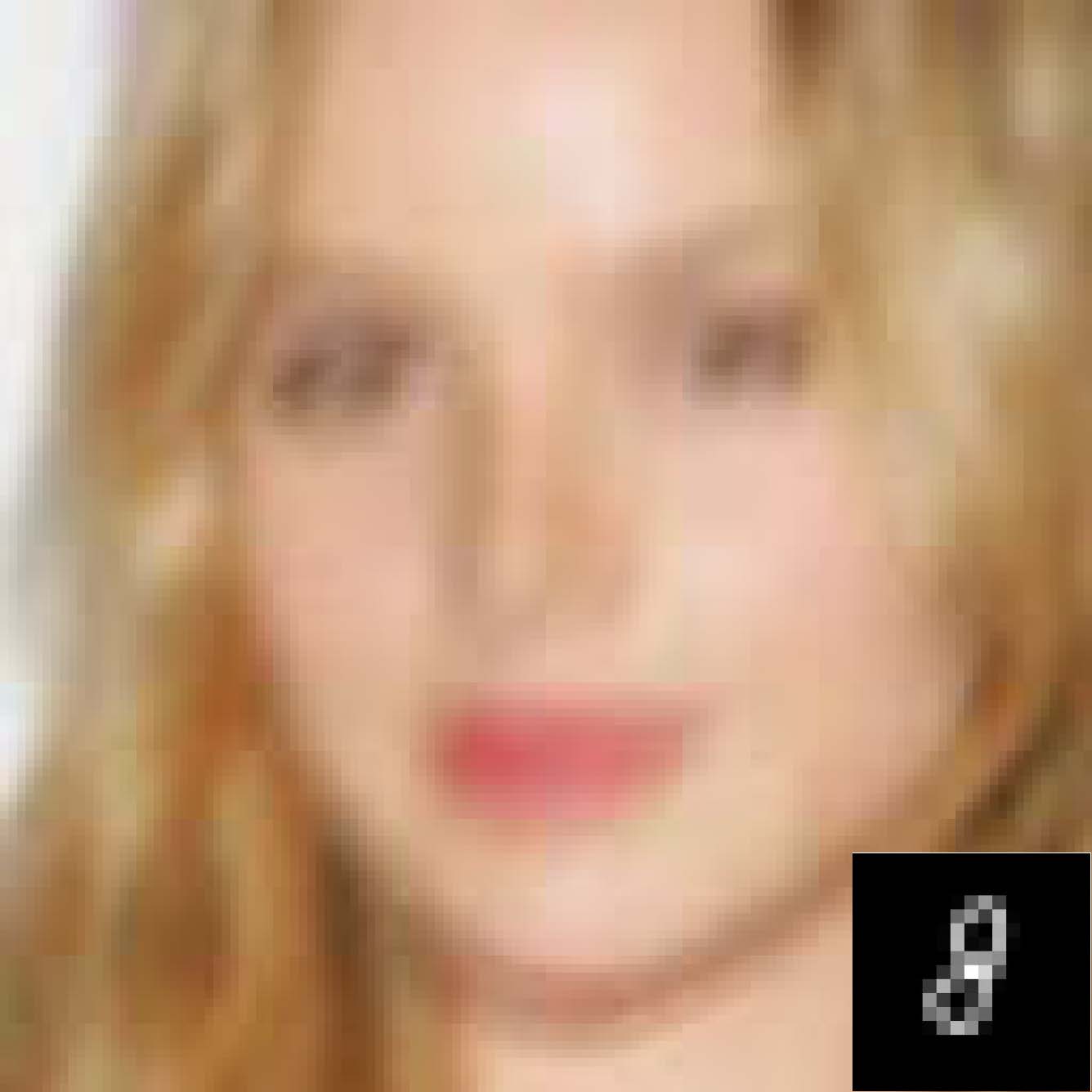} &
            \includegraphics[width=0.19\columnwidth]{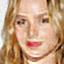} &
            \includegraphics[width=0.19\columnwidth]{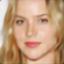} &
            \includegraphics[width=0.19\columnwidth]{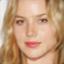}  
            \\
            \includegraphics[width=0.19\columnwidth]{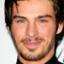} &
            \includegraphics[width=0.19\columnwidth]{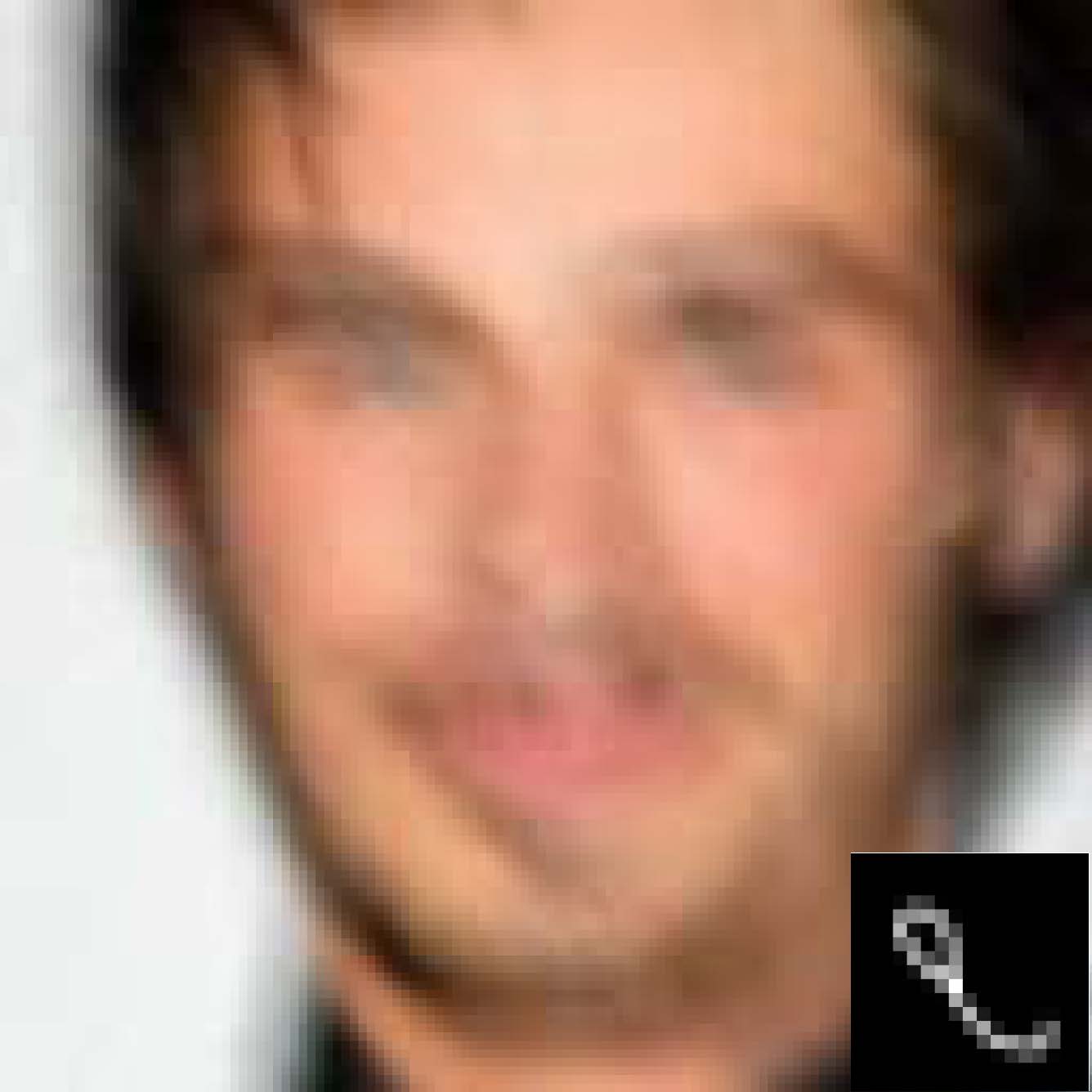} &
            \includegraphics[width=0.19\columnwidth]{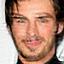} &
            \includegraphics[width=0.19\columnwidth]{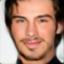} &
            \includegraphics[width=0.19\columnwidth]{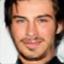} 
            \\
            \includegraphics[width=0.19\columnwidth]{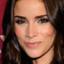} &
            \includegraphics[width=0.19\columnwidth]{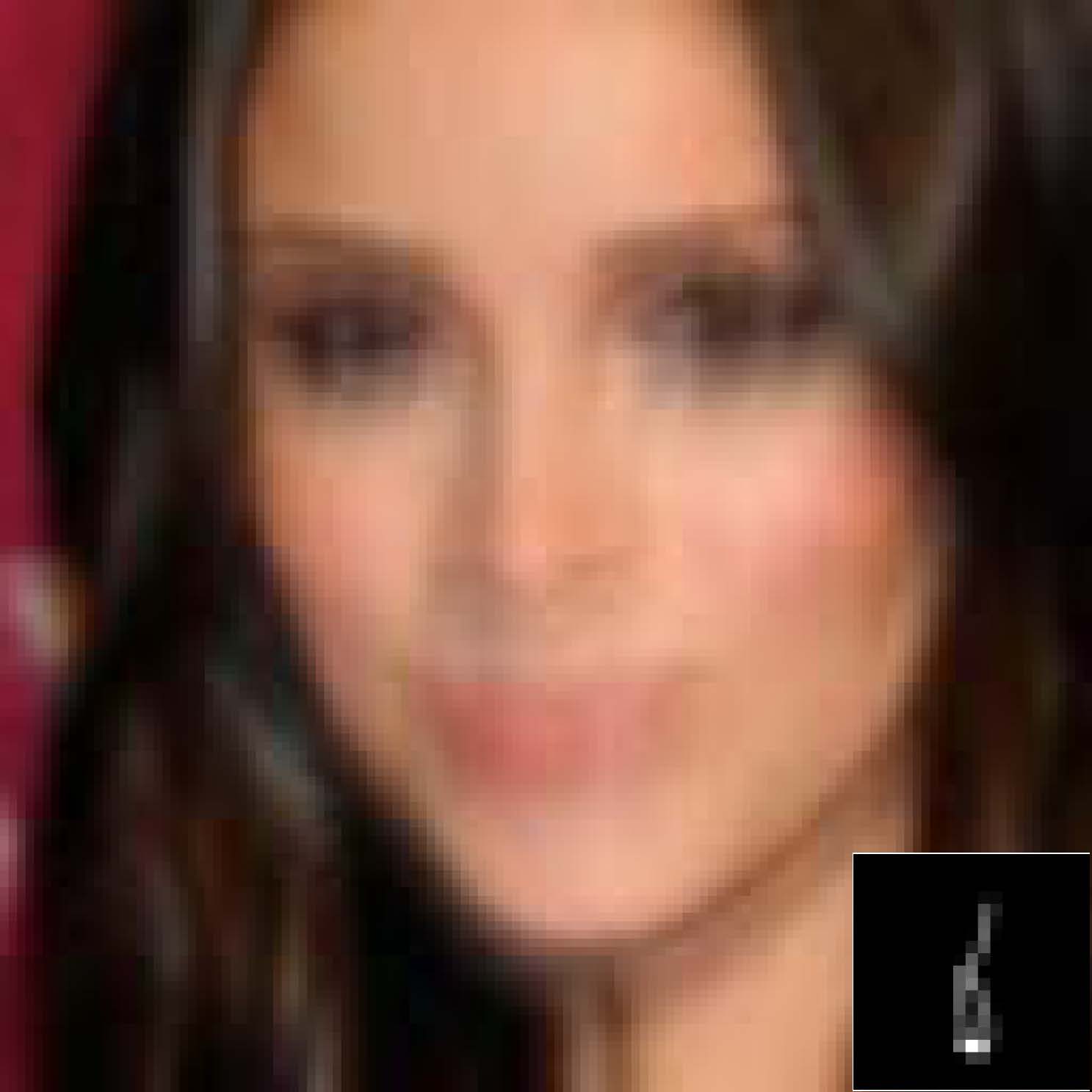} &
            \includegraphics[width=0.19\columnwidth]{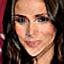} &
            \includegraphics[width=0.19\columnwidth]{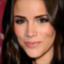} &
            \includegraphics[width=0.19\columnwidth]{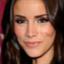} 
            \\
			(a) & (b) & (c) & (d) & (e) \\
		\end{tabular}
	\end{center}
	\vspace{-2mm}
	\caption{\textbf{Visual comparison on face image deblurring}.
		We exploit the semantic information of face images 
		within an end-to-end deep CNN for deblurring.
		(a) Ground truth images
		(b) Blurred images
		(c) \citet{DBLP:conf/eccv/PanHSY14}
		(d) \citet{Ziyi-CVPR-2018}
		(e) Ours.
	}
	\label{fig:teaser}
\end{figure}

To handle blurred images caused by unknown blur kernels, we construct a large face blurred image dataset for training and testing.
We first synthesize random blur kernels by modeling the camera trajectories~\citep{DBLP:conf/eccv/Chakrabarti16,DBLP:conf/bmvc/HradisKZS15}.
Next, we generate blurred face images using the synthesized blur kernels and face images from the Helen~\citep{helen}, CMU PIE~\citep{PIE}, and CelebA~\citep{CelebA} datasets.
We show that the proposed model trained on synthetic images generalizes well to images generated by unseen blur kernels as well as real blurred images.
The proposed method reconstructs better facial details and achieves higher accuracy on face detection and recognition than the state-of-the-art face deblurring approaches~\citep{Ziyi-CVPR-2018,DBLP:conf/eccv/PanHSY14} (see~\figref{teaser}).

In this work, we make the following contributions:
\begin{itemize}
\item We propose a deep multi-scale CNN that exploits global semantic priors and local structural constraints for face image deblurring. 
The proposed local structural loss adaptively adjusts the weights based on the size of each facial component and greatly improves the quantitative and qualitative results.
\item We develop a large-scale blurred face image dataset. 
The training set consists of 130 million blurred images (synthesized from 6,464 face images and 20,000 blur kernels) and the test set has 16,000 blurred images (synthesized from 200 face images and 80 blur kernels).
Our dataset can serve as a common benchmark for training and evaluating face image deblurring.	
\item We demonstrate that the proposed method performs favorably against 
the state-of-the-art deblurring approaches in terms of restoration quality, face detection, recognition and execution speed.
\end{itemize}

%% Introduce our goal and CNN-based method
%
\section{Related Work}
\label{sec:related}

Our work belongs to the single-image blind image deblurring problem, where the blur kernel is unknown.
In this section, we focus our discussion on generic, domain specific, and recent CNN-based image deblurring approaches.

%\vspace{-2mm}
\subsection{Generic image deblurring methods}

The recent advances in single image blind deblurring can be attributed to the development of effective natural image priors, including sparse gradient prior~\citep{DBLP:journals/tog/FergusSHRF06,DBLP:conf/cvpr/LevinWDF09}, normalized sparsity measure~\citep{DBLP:conf/cvpr/KrishnanTF11}, patch prior~\citep{Sun-ICCP-2013}, $L_0$ gradient~\citep{DBLP:conf/cvpr/XuZJ13}, color-line prior~\citep{Lai-CVPR-2015}, low-rank prior~\citep{ren2016image}, self-similarity~\citep{Michaeli-ECCV-2014}, and extreme channel priors~\citep{DBLP:conf/cvpr/PanSP016,yan2017image}.
Recently, a number of approaches learn data fitting functions~\citep{PAN17} or image priors with Markov random fields (MRFs)~\citep{liu18} to recover latent images.
By optimizing the image priors within the MAP framework, those approaches \emph{implicitly} restore strong edges, and therefore, estimate blur kernels and latent sharp images.
However, solving complex non-linear priors involves several optimization steps and thus entails high computational loads.
Edge-selection based methods~\citep{DBLP:journals/tog/ChoL09,DBLP:conf/eccv/XuJ10} use simple priors (e.g., $L_2$ gradients) with image filters (e.g., shock filter) to \emph{explicitly} restore or select strong edges.
In addition, a number of approaches use reference images as guidance for non-blind~\citep{Sun-ECCV-2014} and blind deblurring~\citep{Hacohen-ICCV-2013}.
However, the performance of such methods hinges on the similarity of 
reference images and quality of dense correspondence.

While generic image deblurring methods demonstrate the state-of-the-art performance, face images have different statistical properties than natural scenes.
Fewer edges or structure on face images can be extracted for blur kernel estimation. 
The above-mentioned approaches typically cannot deblur face images well and may 
generate undesired visual artifacts.

Another line of work proposes various motion blur models to handle non-uniform blur~\citep{hirsch2011fast, whyte2012non} and depth variation~\citep{hu2014joint}.
In this work, we focus on face images caused by uniform motion blur.
Our method can also be extended to handle non-uniform blur by synthesizing training data with the non-uniform blur model.

\subsection{Domain specific image deblurring methods}

Several domain specific image deblurring approaches have been developed to handle images from different categories.
As text images usually contain nearly uniform intensity, \citet{DBLP:journals/pami/PanHS017} introduce the $L_0$-regularized priors on both intensity and image gradients for deblurring text images.
To handle extreme cases such as low-light images, \citet{Hu-CVPR-2014} detect the light streaks in images for estimating blur kernels.
\citet{Anwar-ICCV-2015} propose a frequency-domain class-specific prior to restore the band-pass frequency components.
Several recent approaches propose outlier detection methods~\citep{pan2016robust} or robust loss functions~\citep{dong2017blind} to handle images with non-Gaussian noise.

As face images contain fewer textures and edges, existing algorithms based on implicit or explicit edge restoration are less effective.
\citet{DBLP:conf/eccv/PanHSY14} search for similar images from a face dataset and extract reference exemplar contours for blur kernel estimation. 
However, this approach requires manual annotations of the facial contours and involves computationally expensive optimization within the MAP framework.
In contrast, we train an end-to-end deep CNN to bypass the blur kernel estimation step, without requiring any reference images or manual annotations for face deblurring.

\subsection{CNN-based image deblurring methods}
Deep CNNs have been adopted for several image restoration tasks, such as denoising~\citep{Mao-NIPS-2016}, JPEG deblocking~\citep{Dong-ICCV-2015}, dehazing~\citep{ren2016single} and super-resolution~\citep{VDSR,LapSRN}.
Several methods apply deep CNNs for image deblurring in different aspects, including non-blind deconvolution~\citep{Schuler-CVPR-2013,DBLP:conf/nips/XuRLJ14,Zhang_2017_CVPR}, blur kernel estimation~\citep{DBLP:conf/cvpr/SunCXP15,DBLP:journals/pami/SchulerHHS16,DBLP:conf/eccv/Chakrabarti16}, and dynamic scene deblurring~\citep{Nah_2017_CVPR,tao2018scale}.
%
%Despite the computational efficiency, CNN-based methods do not perform as well as
the state-of-the-art MAP-based approaches, especially in the presence of large motion.
Several approaches embed deep CNNs into the conventional MAP-based framework by learning discriminative image priors~\citep{li2018learning} or predicting sharp edges~\citep{DBLP:journals/tip/XuPZY18} to achieve the state-of-the-art performance.
More recently, \citet{NimishaSR17} and \citet{deblurgan} train generative adversarial networks for blind motion deblurring.

A number of methods train end-to-end networks to handle class-specific images, e.g., texts~\citep{DBLP:conf/bmvc/HradisKZS15} and faces~\citep{Jin18,ChrysosFZ19}.
\citet{Xu-ICCV-2017} train generative adversarial networks to jointly deblur and super-resolve low-resolution blurred face and text images, which are typically degraded by Gaussian-like blur kernels.
A few face deblurring methods~\citep{Jin18, ChrysosFZ19} based on generic CNNs have recently been developed. 
Although there are some implementation differences in network architectures and loss functions (e.g., the model of \citep{Jin18} is lightweight conditions), these methods do not explore face-related prior information to help the deblurring process.
In this work, we focus on deblurring face images affected by complex motion blur.
We exploit global and local semantic cues as well as the perceptual~\citep{Johnson-ECCV-2014} and adversarial~\citep{GAN} losses to restore photo-realistic face images with fine details.

\begin{figure}
	\centering
	\footnotesize
	\includegraphics[height=0.15\textwidth]{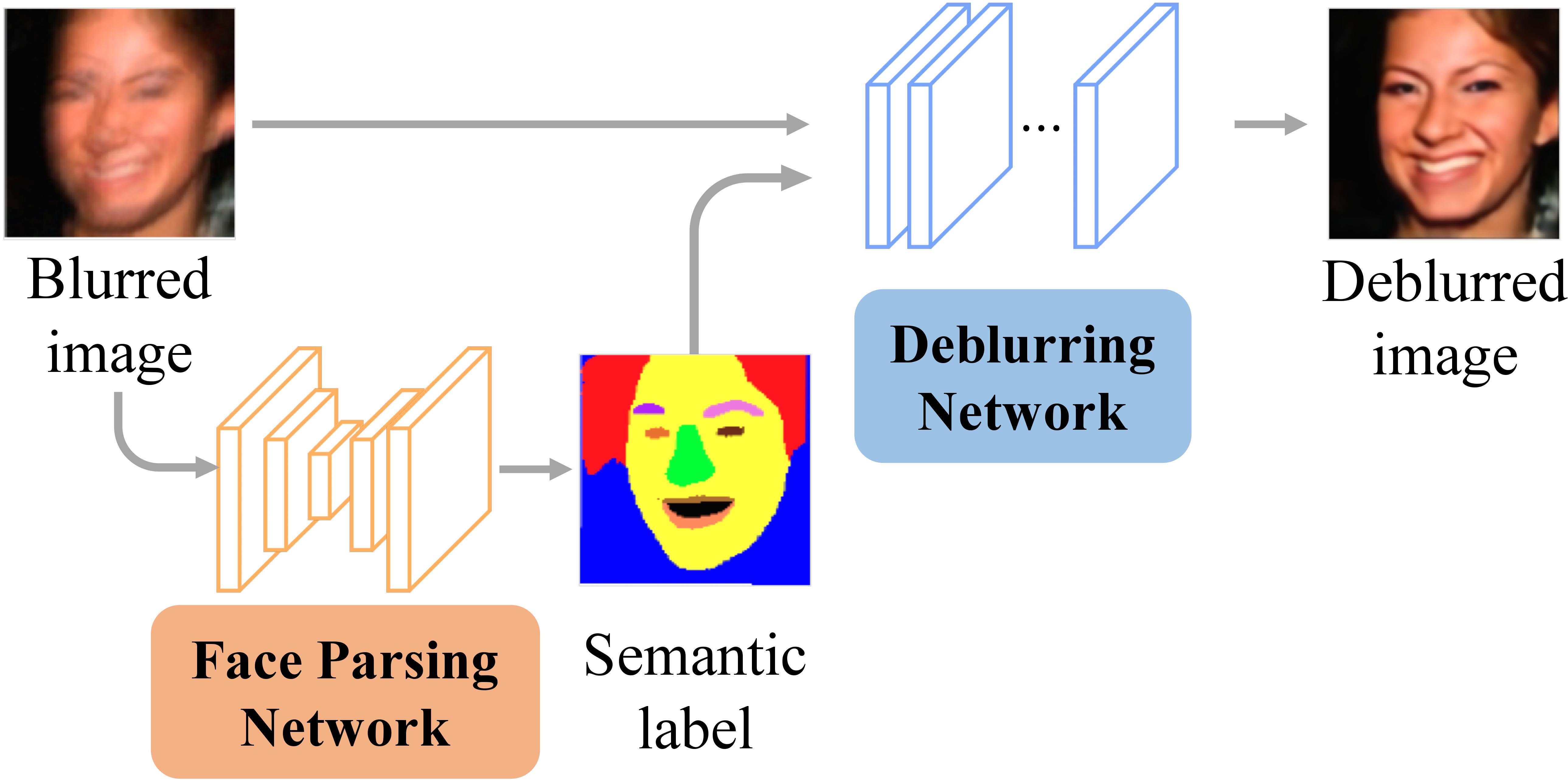} \\
	(a) \citet{Ziyi-CVPR-2018} 
	\\ %\vspace{2mm}
	\includegraphics[width=\columnwidth]{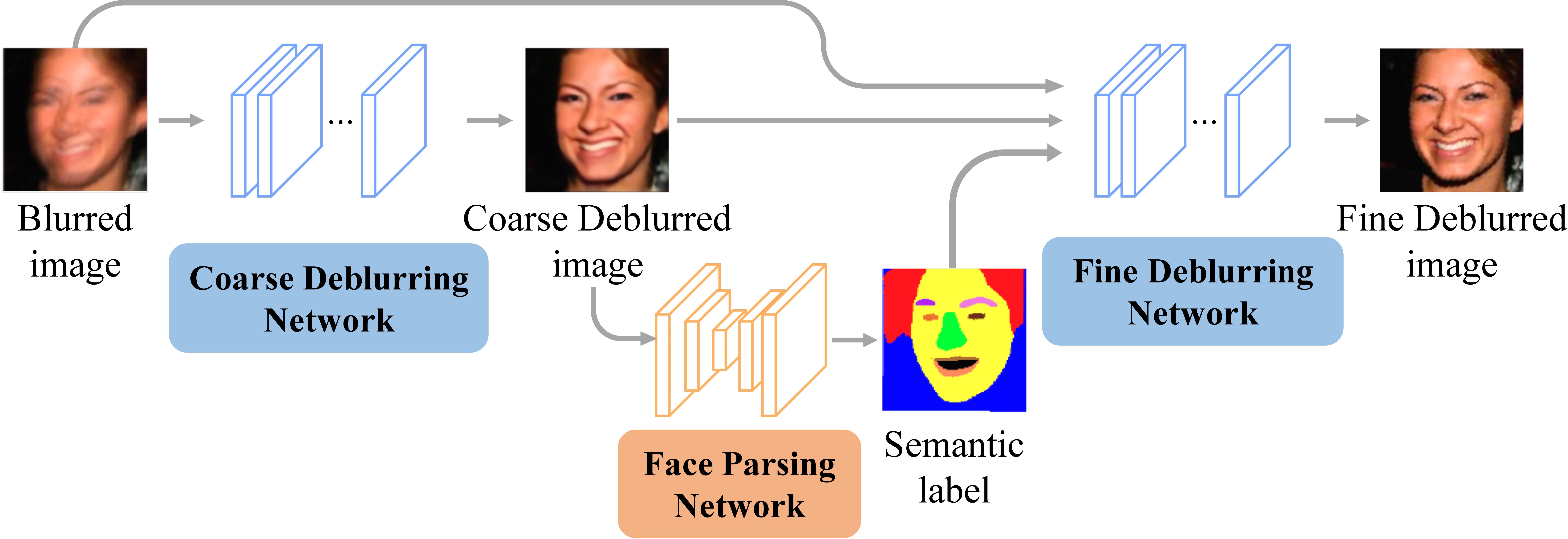} \\
	(b) Ours \\
	\caption{
		\textbf{Overview of the proposed model.}
		The state-of-the-art method~\citep{Ziyi-CVPR-2018} extracts the semantic labels from a \emph{blurred} image, while we obtain the semantic labels from a \emph{coarse deblurred} image.
		The coarse deblurring network reduces the motion blur from the input image and leads to more accurate face parsing results.
		}
	\label{fig:overview}
\end{figure}

\section{Semantic Face Deblurring}
\label{sec:algorithm}
In this section, we first give an overview of the proposed face deblurring method.
We then describe the design methodology of the network architecture, loss functions, and implementation details.

\vspace{-3mm}
\subsection{Overview}
We aim to utilize the face semantic cues to deblur face images.
In our preliminary work~\citep{Ziyi-CVPR-2018}, we first apply a face parsing network to extract semantic labels from the input blurred image and then adopt a deblurring network 
for restoration. 
We also propose a local structural loss to enforce additional weights on important facial components to recover fine details.
However, the labels extracted from the blurred images may be erroneous due to 
severe motion blur.
In this work, we make the following improvements:
\begin{itemize}
\item We first construct a coarse deblurring network to reduce the blur in the input image.
The face parsing network then extracts semantic labels from the coarse deblurred image.
Finally, the fine deblurring network restores a clear face image from the given blurred input image, coarse deblurred image, and corresponding semantic label maps.
\item Instead of using a fixed weight for all key components, we propose an adaptive local structural loss which adjusts the weight based on the size of each facial component and restores more fine details.
\end{itemize}
\figref{overview} shows the differences between the method of~\citet{Ziyi-CVPR-2018} and proposed model.

\begin{figure*}
	\centering
	\footnotesize
	\includegraphics[width=0.9\textwidth]{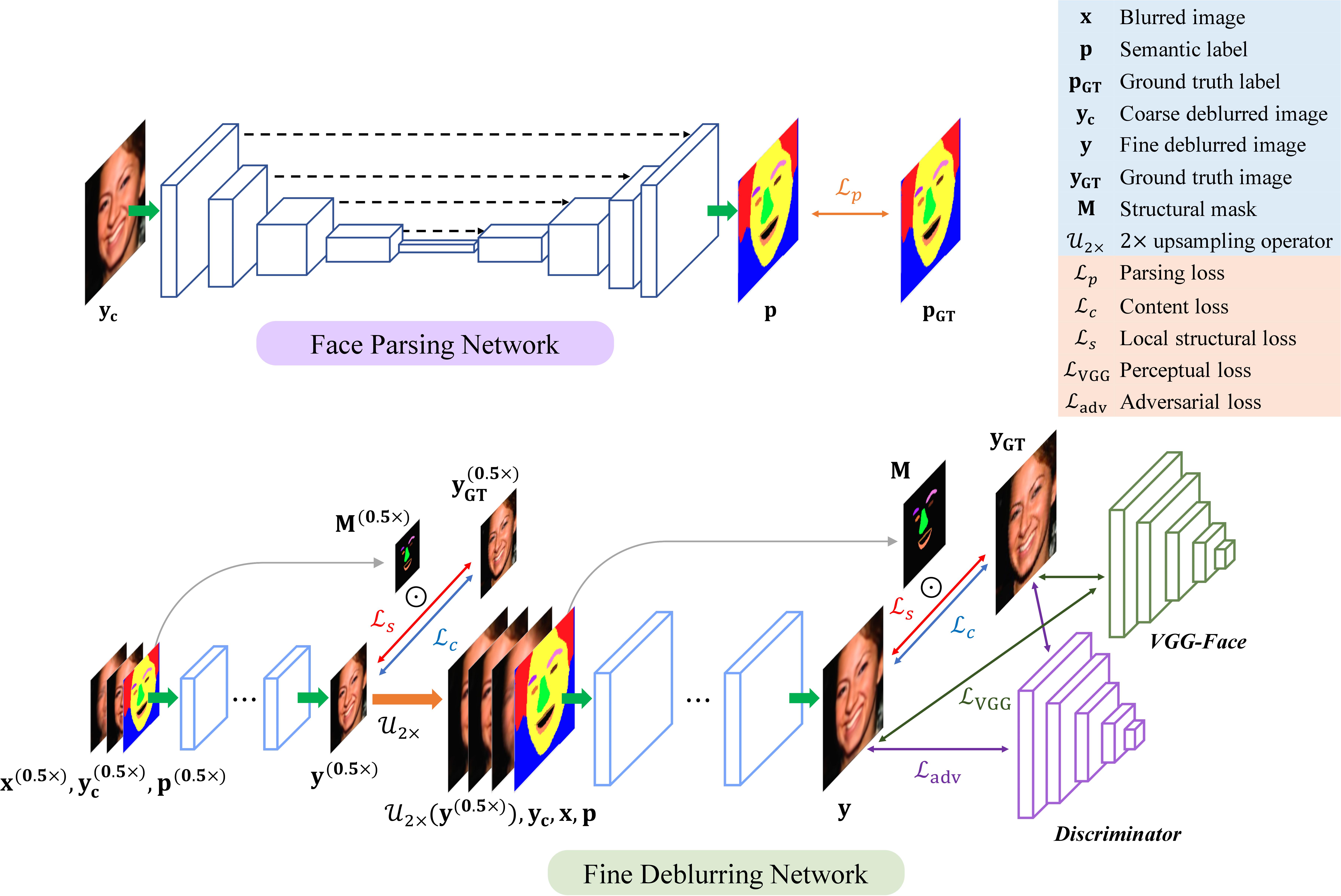}
%	\vspace{1mm}
	\caption{
		\textbf{Architecture of the proposed model.}
		The face parsing network is an encoder-decoder architecture with skip connections from the encoder to the decoder.
		The fine deblurring network has two scales.
		The first scale generates a deblurred image with $0.5\times$ spatial resolution, and the second scale generates a full-resolution deblurred image.
		Each scale of the deblurring network receives the supervision from the pixel-wise content loss and local structural loss.
		In addition, we impose the perceptual and adversarial losses at the output of the second scale.
		The coarse deblurring network has a similar architecture to the fine deblurring network but without taking the semantic label as input and only receiving supervision from the content loss.
		}
	\label{fig:network}
	%\vspace{1mm}
\end{figure*}

\subsection{Network architecture}
\label{sec:deblur_network}

Given a blurred face image $\mathbf{x} \in \mathbb{R}^{H \times W \times 3}$ where $H$ and $W$ denotes the height and width of the image, our goal is to recover a clear and sharp face image $\mathbf{y}$ which is as similar as the ground truth image $\mathbf{y}_{\text{GT}}$. 
To this end, we train an end-to-end deep CNN to deblur the face images efficiently.
The proposed face deblurring model consists of three sub-networks: a coarse deblurring network $\mathcal{G}_c$, a face parsing network $\mathcal{P}$, and a fine deblurring network $\mathcal{G}_f$.

\Paragraph{Coarse deblurring network.}
To reduce the influence of motion blur on the face parsing, we first use a network to obtain a coarse deblurred image $\mathbf{y}_c$:
\begin{equation}
    \mathbf{y}_c = \mathcal{G}_c(\mathbf{x}).
    \label{eq:coarse_deblur}
\end{equation}
We use a multi-scale network similar to the model of~\citet{Nah_2017_CVPR}, but with several differences.
First, as face images typically have smaller spatial resolutions (e.g., $128 \times 128$ or less), we use only 2 scales instead of 3 scales for natural images in~\citep{Nah_2017_CVPR}.
Second, we use fewer ResBlocks (reduce from 19 to 6) and a larger filter size ($11 \times 11$) at the first convolutional layer to increase the receptive field.
The first scale takes as input the $2\times$ downsampled blurred image $\mathbf{x}^{(0.5\times)}$ (3 channels) and generates a deblurred image $\mathbf{y}_c^{(0.5\times)}$.
The input to the second scale contains the blurred image $\mathbf{x}$ (3 channels) and the upsampled deblurred image from the first scale $\mathcal{U}_{2\times}(\mathbf{y}_c^{(0.5\times)})$ (3 channels), where $\mathcal{U}_{2\times}$ is $2\times$ bicubic upsampling operator.
The output image from the second scale is the coarse deblurred result $\mathbf{y}_c$.

\Paragraph{Face parsing network.}
We use an encoder-decoder architecture with skip connections as our face parsing network.
The face parsing network takes the coarse deblurred image as input and generates the probability map of face semantic labels $\mathbf{p} \in \mathbb{R}^{H \times W \times K}$:
\begin{equation}
    \mathbf{p} = \mathcal{P}(\mathbf{y}_c), 
    \label{eq:parsing}
\end{equation}
where $K$ is the number of semantic classes.
The semantic probabilities encode the essential appearance information and approximate locations of the facial components (e.g., eyes, noses and mouths) and serve as a strong global prior for reconstructing the deblurred face image.
We extract $K = 14$ semantic labels (see~\tabref{parsing_accuracy}) for each input image.

\Paragraph{Fine deblurring network.}
The fine deblurring network has a similar architecture to the coarse deblurring network.
In addition, we take as input the blurred image $\mathbf{x}$, coarse deblurred image $\mathbf{y}_c$, as well as the semantic probability maps $\mathbf{p}$ to recover a clear face image $\mathbf{y}$:
\begin{equation}
    \mathbf{y} = \mathcal{G}_f(\mathbf{x}, \mathbf{y}_c, \mathbf{p}).
    \label{eq:fine_deblur}
\end{equation}

Our fine deblurring network also has a similar two-scale structure to the coarse deblurring network.
The input to the first scale includes the $2\times$ downsampled blurred image $\mathbf{x}^{(0.5\times)}$ (3 channels), $2\times$ downsampled coarse deblurred image $\mathbf{y}_c^{(0.5\times)}$ (3 channels), and the $2\times$ downsampled semantic probability maps $\mathbf{p}^{(0.5\times)}$ (11 channels), resulting in a 17-channel input feature.
The input to the second scale includes the blurred image $\mathbf{x}$ (3 channels), coarse deblurred image $\mathbf{y}_c$ (3 channels), the upsampled deblurred image from the first scale $\mathcal{U}_{2\times}(\mathbf{y}^{(0.5\times)})$ (3 channels), and the semantic probability maps $\mathbf{p}^{(0.5\times)}$ (11 channels), resulting in a 20-channel input feature.
The output image from the second scale of the fine deblurring network is the final deblurred result $\mathbf{y}$.
\figref{network} shows an overview of our face parsing and deblurring network.

\subsection{Loss functions}
\label{sec:loss}
We train the parsing network using a cross-entropy loss and optimize the deblurring networks with a pixel-wise content loss and the proposed adaptive local structural loss.
As pixel-wise $L_2$ or $L_1$ loss functions typically lead to overly-smooth results, we further introduce a perceptual loss~\citep{Johnson-ECCV-2014} and an adversarial loss~\citep{GAN} to optimize our deblurring network and generate photo-realistic deblurred results.

\Paragraph{Parsing loss.}
We adopt a multi-class cross-entropy loss function to optimize the face parsing network:
\begin{equation}
	\mathcal{L}_p = - \sum_{k=1}^K \mathbf{p}_{\text{GT}}^{(k)} \log(\mathbf{p}^{(k)}), 
	\label{eq:parsing_loss}
\end{equation}
where $\mathbf{p}_{\text{GT}}^{(k)}$ is the ground truth semantic label for the $k^{\text{th}}$ class.

\Paragraph{Content loss.}
We adopt the pixel-wise $L_1$ robust function as the content loss of the coarse and fine deblurring networks:
\begin{equation}
	\mathcal{L}_c = 
	    \left\| \mathbf{y}_c - \mathbf{y}_{\text{GT}} \right\|_1 + 
	    \left\| \mathbf{y} - \mathbf{y}_{\text{GT}} \right\|_1.
	\label{eq:content_loss}
\end{equation}

\Paragraph{Adaptive local structural loss.}
While the content loss~\eqnref{content_loss} enforces a holistic supervision from the ground truth clear image, the key components (e.g., eyes, lips, and mouths) on faces may be easily ignored as they are typically thin and small.
Solely minimizing the content loss on the whole face image cannot guarantee to restore the fine details.
Thus, we propose to impose a local structural loss on facial key components:
\begin{equation}
	\mathcal{L}_s = \sum_{k = 1}^K w_k \left\| \mathbf{M}_k \odot \mathbf{y} - \mathbf{M}_k \odot \mathbf{y}_{GT} \right\|_1,
	\label{eq:local_loss}
\end{equation}
where $w_k$ is the weight of each component and $\mathbf{M}_k$ denotes the structural mask of the $k^{\text{th}}$ component (extracted from the semantic label $\mathbf{p}$).
We apply the local structural losses on eight important components, including left eye, right eye, left eyebrow, right eyebrow, nose, upper lip, lower lip, and teeth, to enhance the local details.
We do not apply the local structural loss on textureless regions, such as hair and skin.
The local structural losses enforce the deblurring network to restore more details with fewer artifacts on the face images.

In our preliminary work~\citep{Ziyi-CVPR-2018}, we adopt an \emph{equal} weight for all the selected components, i.e., $w_k = 1$, $\forall k = 1, \cdots, K$.
However, tiny components (e.g., eyes) may not be well reconstructed when optimizing the network.
In this work, we propose an \emph{adaptive} weighting mechanism based on the size of each component:
\begin{equation}
    w_k = c / A_k, 
	\label{eq:adaptive_weight}
\end{equation}
where $c$ is a constant and $A_k$ is the size of the  $k^{\text{th}}$ component.
The adaptive local structural loss enforces larger weights on small components and thus  helps recover facial details.

\Paragraph{Perceptual loss.}
The perceptual loss has been adopted in style transfer~\citep{DBLP:conf/nips/GatysEB15,Johnson-ECCV-2014}, image super-resolution~\citep{Ledig_2017_CVPR} and image synthesis~\citep{Chen-ICCV-2017,wang2017high}.
The perceptual loss aims to measure the similarity in the high dimensional feature space of a pre-trained classification network (e.g., VGG16~\citep{conf/ICLR/Simonyan15}).
Given the input image $x$, we denote $\phi_l(x)$ as the activation at the $l$-th layer of the loss network $\phi$.
The perceptual loss is then defined as:
\begin{equation}
	\mathcal{L}_{\text{VGG}} = \sum_{l} \left\| \phi_l(\mathbf{y}) - \phi_l(\mathbf{y}_{\text{GT}}) \right\|_1.
	\label{eq:perceptual_loss}
\end{equation}
We compute the perceptual loss on the \texttt{pool2} and \texttt{pool5} layers of the pre-trained VGG-Face~\citep{VGG-Face}.

\Paragraph{Adversarial loss.}
The adversarial training framework has been effectively applied to synthesize realistic images~\citep{GAN,Ledig_2017_CVPR,Nah_2017_CVPR}.
We treat our fine deblurring network as the generator and construct a discriminator based on the DCGAN~\citep{DCGAN} model.
The goal of the discriminator $\mathcal{D}$ is to distinguish the real image from the output of the generator.
The generator $\mathcal{G}$ aims to generate images as real as possible to fool the discriminator.
The adversarial training is formulated as solving the following min-max problem:
\begin{align}
	\underset{\mathcal{G}}{\mathop{\min }}\,
	\underset{\mathcal{D}}{\mathop{\max }}\,
	\mathbb{E} \left[ \log \mathcal{D}(\mathbf{y}_{\text{GT}}) \right]
	+ \mathbb{E} \left[ \log (1-\mathcal{D}( \mathbf{y} )) \right].
	\label{eq:adversarial_loss}
\end{align}
When updating the generator, the adversarial loss is:
\begin{equation}
	\mathcal{L}_{\text{adv}} = - \log \mathcal{D}( \mathbf{y} ).
	\label{eq:adv_G}
\end{equation}
Our discriminator takes an input image with of $128 \times 128$ pixels and has 6 strided convolutional layers followed by the ReLU activation function.
In the last layer, we use the sigmoid function to output a single scalar 
as the probability of being a real image.
Similar to existing image super-resolution~\citep{Ledig_2017_CVPR} and motion deblurring~\citep{deblurgan, Nah_2017_CVPR} methods, 
the generator of the proposed model does not take a noise vector as input.

\Paragraph{Overall loss function.}
The overall loss function for training our face deblurring model is:
\begin{equation}
	\mathcal{L} = \mathcal{L}_c + \lambda_s \mathcal{L}_s + \lambda_p \mathcal{L}_p + \lambda_{\text{VGG}} \mathcal{L}_{\text{VGG}} + \lambda_{\text{adv}} \mathcal{L}_{\text{adv}},
	\label{eq:overall_loss}
\end{equation}
where $\lambda_s$, $\lambda_p$, $\lambda_{\text{VGG}}$, and $\lambda_{\text{adv}}$ are the weights to balance the local structural losses, parsing loss, perceptual loss and adversarial loss, respectively.
In this work, we empirically set the weights to $\lambda_s = 50$, $\lambda_p = 1e^{-4}$, $\lambda_{\text{VGG}} = 1e^{-5}$, $\lambda_{\text{adv}} = 5e^{-5}$, and the constant $c = 1$ in~\eqnref{adaptive_weight}.
We adopt the content and local structural losses at all scales of the deblurring network while only apply the perceptual and adversarial losses to the final output image, i.e., the output of second scale from the fine deblurring network.

\subsection{Training strategy}
\label{sec:training_strategy}
As our model consists of three sub-networks, it is difficult to jointly optimize the whole model simultaneously.
We adopt the following progressive training strategy:
\begin{enumerate}
\item We first train the coarse deblurring network $\mathcal{G}_c$ using the content loss~\eqnref{content_loss} on the coarse deblurred image for 200,000 iterations.
\item We then fix $\mathcal{G}_c$ and train the face parsing network $\mathcal{P}$ using the parsing loss~\eqnref{parsing_loss} for 60,000 iterations. 
\item Next, we fix both $\mathcal{G}_c$ and $\mathcal{P}$ and train the fine deblurring network $\mathcal{G}_f$ using the content loss~\eqnref{content_loss}, local structural loss~\eqnref{local_loss}, perceptual loss~\eqnref{perceptual_loss} and adversarial loss~\eqnref{adversarial_loss} for 200,000 iterations.
\item Finally, we jointly optimize all three sub-networks by minimizing the overall loss~\eqnref{overall_loss} for 100,000 iterations.
\end{enumerate}
We demonstrate that such a progressive training strategy can achieve better performance then jointly training the whole model from scratch in \secref{analysis}.

\subsection{Implementation details}
Both the coarse and fine deblurring networks have two scales, where each scale has 6 ResBlock~\citep{he2016deep} (include two convolutional layers and one activation layer) and 18 convolutional layers.
The first convolutional layer at each scale has a kernel size of $11 \times 11$, while all other convolutional layers have a kernel size of $5 \times 5$ and 64 channels.
The upsampling layer uses a $4 \times 4$ transposed convolutional layer to upsample the image by $2\times$.
We use the ReLU as the activation function and do not use any normalization layer (e.g., batch normalization).

We implement our network using the MatConvNet toolbox~\citep{matconvnet}.
We use a batch size of 16 and set the learning rate to $5e^{-6}$ for the parsing network and $4e^{-5}$ for the coarse and fine deblurring networks.
During the training process, we apply the following data augmentation: (1) random scaling between $[0.9, 1.1]\times$, (2) random horizontal and vertical shifting within 12 pixels, and (3) random rotating within $\pm 30^{\circ}$.
The whole training process takes about 5 days on an NVIDIA Titan X GPU card.

\begin{table}
	\centering
	\footnotesize
	\caption{
	    \textbf{Summary of our face deblurring dataset.}
	    We collect clear face images from the Helen~\citep{helen}, CMU PIE~\citep{PIE}, and CelebA~\citep{CelebA} datasets and synthesize blur kernels for generating blurred face images.
	}
	\label{tab:dataset}
	%\vspace{1mm}
	\begin{tabular}{c|ccc|c|c}
		\toprule
		         & \multicolumn{3}{c|}{Clear images} & Blur & Blurred \\
		         & Helen & CMU PIE & CelebA & Kernels & Images \\
		\midrule
		Training & 2000 & 2164 & 2300 & 20,000 & 130 M \\
		Testing  &  100 &  -   &  100 & 80 & 16,000 \\
		\bottomrule
	\end{tabular}
\end{table}

\subsection{Face deblurring datasets}
We collect clear face images from the Helen~\citep{helen}, CMU PIE~\citep{PIE}, and CelebA~\citep{CelebA} datasets.
We align all the face images by first detecting the facial landmarks using the method of~\citet{DBLP:conf/cvpr/SunWT13} and warping the images based on the aligned landmarks~\citep{DBLP:conf/cvpr/KaeSLL13}.
The motion blur kernels are synthesized by modeling random 3D camera trajectories~\citep{DBLP:journals/tip/BoracchiF12}.
We generate blur kernels with 8 different sizes (from $13 \times 13$ to $27 \times 27$).
By convolving the clear images with blur kernels and adding Gaussian noise with $\sigma = 0.01$, we obtain 130 million blurred images for training and 16,000 blurred images for testing.
\tabref{dataset} summarizes the number of clear face images, motion blur kernels, and synthesized blurred images in the training and testing sets.
We note that the 20,000 blur kernels used to generate training images are different from the 80 blur kernels used in the test set.
Both the clear faces images and blur kernels are disjoint in the training and testing sets.

\section{Analysis and Discussions}
\label{sec:analysis}

In this section, we first demonstrate the effectiveness of using semantic parsing labels for face image deblurring.
We then conduct ablation studies to analyze the contribution of each sub-network and loss function.

\subsection{Effect of semantic parsing}
Our key idea is to utilize the face semantic labels as prior information to facilitate the face deblurring.
We first validate the idea by using the \emph{ground truth} semantic labels as an additional input to our deblurring network.
Since only the Helen dataset contains ground truth face labels, we first train a face parsing network using the clear images and ground truth from the Helen dataset.
We then use this face parsing network to generate labels for the clear images in the CMU PIE and CelebA datasets, which are treated as the \emph{pseudo ground-truth} labels to train the proposed face parsing network for deblurring.

We train a baseline model $\mathcal{G}$ using the coarse deblurring network, which does not take any semantic information as input and does not adopt the local structural loss.
Then, we concatenate the ground truth semantic labels $\mathbf{p}_{\text{GT}}$ with the blurred image as input to the baseline model.
We evaluate the PSNR and SSIM on the Helen test set and present the results in~\tabref{analysis_parsing}.
The model with prior knowledge from the ground truth labels ($2^{nd}$ row) significantly outperforms the baseline model ($1^{st}$ row), which demonstrates the effect of semantic labels on deblurring face images.

In~\citet{Ziyi-CVPR-2018}, the semantic labels are extracted from the \emph{blurred} images.
While the parsing network $\mathcal{P}$ is fine-tuned on blurred images for performance gain, the semantic labels of some small components (e.g., eyebrows, lips, and teeth) may not be accurate enough when the input image suffers from large motion blur.
In the proposed method, we first apply a coarse deblurring network $\mathcal{G}_c$ to reduce the motion blur and recover a rough structure of the input face image.
We then fine-tune the parsing network $\mathcal{P}$ on the coarse deblurred images and train the fine deblurring network $\mathcal{G}_f$ using the labels extracted from the \emph{coarse deblurred} images.
\tabref{analysis_parsing} shows the performance difference between the method of~\citet{Ziyi-CVPR-2018} ($3^{rd}$ row) and the proposed model ($4^{th}$ row).
The proposed method achieves higher label accuracy and obtains better deblurring results. 
We note that we only use the content loss~\eqnref{content_loss} to train the models in~\tabref{analysis_parsing}.
We also fix the coarse deblurring network and parsing network when training the fine deblurring network to rule out the influence of model parameters.

\begin{figure}
	\footnotesize
	\renewcommand{\tabcolsep}{1pt} % adjust horizontal space
	\renewcommand{\arraystretch}{0.5} % adjust vertical space
	\begin{center}
		\begin{tabular}{ccccc}
			\includegraphics[width=0.092\textwidth]{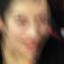} &
			\includegraphics[width=0.092\textwidth]{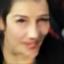} &
			\includegraphics[width=0.092\textwidth]{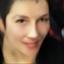} &
            \includegraphics[width=0.092\textwidth]{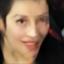} &
			\includegraphics[width=0.092\textwidth]{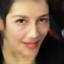} \\
			\includegraphics[width=0.092\textwidth]{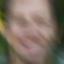} &
            \includegraphics[width=0.092\textwidth]{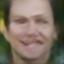} &
            \includegraphics[width=0.092\textwidth]{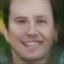} &
            \includegraphics[width=0.092\textwidth]{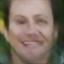} &
			\includegraphics[width=0.092\textwidth]{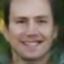} \\
            (a) & (b) & (c) & (d) & (e)
		\end{tabular}
	\end{center}
	\vspace{-2mm}
	\caption{
		\textbf{Deblurred results using different semantic labels.} 
		(a) Blurred images
        (b) Baseline (w/o semantic labels)
        (c) Using ground truth semantic labels
		(d) Using labels from blurred images
		(e) Using labels from coarse deblurred images
	}
	\label{fig:parsing_deblur}
	%\vspace{3mm}
\end{figure}

\begin{figure}
	\footnotesize
	\renewcommand{\tabcolsep}{1pt} % adjust horizontal space
	\renewcommand{\arraystretch}{0.5} % adjust vertical space
	\begin{center}
		\begin{tabular}{ccccc}
            \includegraphics[width=0.19\columnwidth]{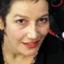} &
            \includegraphics[width=0.19\columnwidth]{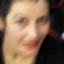} &
            \includegraphics[width=0.19\columnwidth]{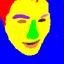} &
            \includegraphics[width=0.19\columnwidth]{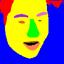} &
            \includegraphics[width=0.19\columnwidth]{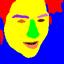} \\
            \includegraphics[width=0.19\columnwidth]{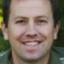} &
            \includegraphics[width=0.19\columnwidth]{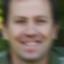} &
            \includegraphics[width=0.19\columnwidth]{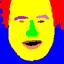} &
            \includegraphics[width=0.19\columnwidth]{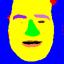} &
            \includegraphics[width=0.19\columnwidth]{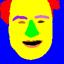} \\
            (a) & (b) & (c) & (d) & (e)\\
		\end{tabular}
	\end{center}
	\vspace{-2mm}
	\caption{
		\textbf{Labeling results of face parsing network.}
		(a) Ground truth clear images
		(b) Input blurred images
		(c) Ground truth semantic labels
		(d) Semantic labels from blurred images
		(e) Semantic labels from coarse deblurred images
	}
	\label{fig:parsing_label}
	%\vspace{-4mm}
\end{figure}

\figref{parsing_deblur} shows the deblurred images by the models listed in~\tabref{analysis_parsing}.
\tabref{parsing_accuracy} shows the parsing accuracy (in terms of the F-score) of each component, and \figref{parsing_label} visualizes the parsing results.
It is clear that more accurate semantic labels provide stronger priors to achieve better deblurring results.

\begin{table}
	\centering
	\footnotesize
	\caption{
	    \textbf{Effect of semantic labels on face image deblurring.}
	    We evaluate the average labeling accuracy (i.e., F-score), PSNR and SSIM of the deblurred images on the Helen dataset.
	}
	\label{tab:analysis_parsing}
	%\vspace{1mm}
	\begin{tabular}{c|ccc}
		\toprule
	    Model
	    & PSNR & SSIM & F-score \\
		\midrule
	    $\mathcal{G}$ &
        24.85 & 0.849 & N.A. \\
	    $\mathbf{p}_{\text{GT}} + \mathcal{G}$ &
	    25.85 & 0.866 & 1.0 \\
		$\mathcal{P}$ (fixed) + $\mathcal{G}$ &
		25.32 & 0.857  & 0.615 \\
		$\mathcal{G}_c$ (fixed) + $\mathcal{P}$ (fixed) + $\mathcal{G}_f$ & 
		25.48 & 0.860  & 0.628 \\
		\bottomrule
	\end{tabular}
	%\vspace{3mm}
\end{table}

\begin{table}
	\centering
	\footnotesize
	\caption{\textbf{Performance of face parsing network.}
		We measure the F-score for each facial component on the Helen dataset.
	}
	\label{tab:parsing_accuracy}
	%\vspace{1mm}
	\begin{tabular}{c|C{1.25cm}C{1.25cm}C{1.25cm}}
		\toprule
		Input image     & Clear & Blurred & Deblurred \\
		\midrule
		face            & 0.915 & 0.886 & 0.881 \\
		left eyebrow    & 0.733 & 0.587 & 0.640 \\
		right eyebrow   & 0.721 & 0.596 & 0.642 \\
		left eye        & 0.741 & 0.679 & 0.655 \\
		right eye       & 0.774 & 0.601 & 0.665 \\
		nose            & 0.899 & 0.864 & 0.872 \\
		upper lip       & 0.653 & 0.477 & 0.502 \\
		lower lip       & 0.733 & 0.632 & 0.625 \\
		teeth           & 0.397 & 0.325 & 0.337 \\
		hair            & 0.566 & 0.499 & 0.466 \\
		\midrule
		average         & 0.713 & 0.615 & 0.628 \\
		\bottomrule
	\end{tabular}
	%\vspace{3mm}
\end{table}

\subsection{Ablation study}
In this section, we analyze the contribution of loss functions, training strategy, and several design choices of the proposed model, including the kernel size, multi-stage deblurring, and effective range of hyper-parameters.

\begin{figure}
    \footnotesize
    \renewcommand{\tabcolsep}{1pt} % adjust horizontal space
    \renewcommand{\arraystretch}{0.5} % adjust vertical space
    \centering
    \begin{tabular}{ccccc}
        \includegraphics[width=0.19\columnwidth]{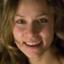} &
        \includegraphics[width=0.19\columnwidth]{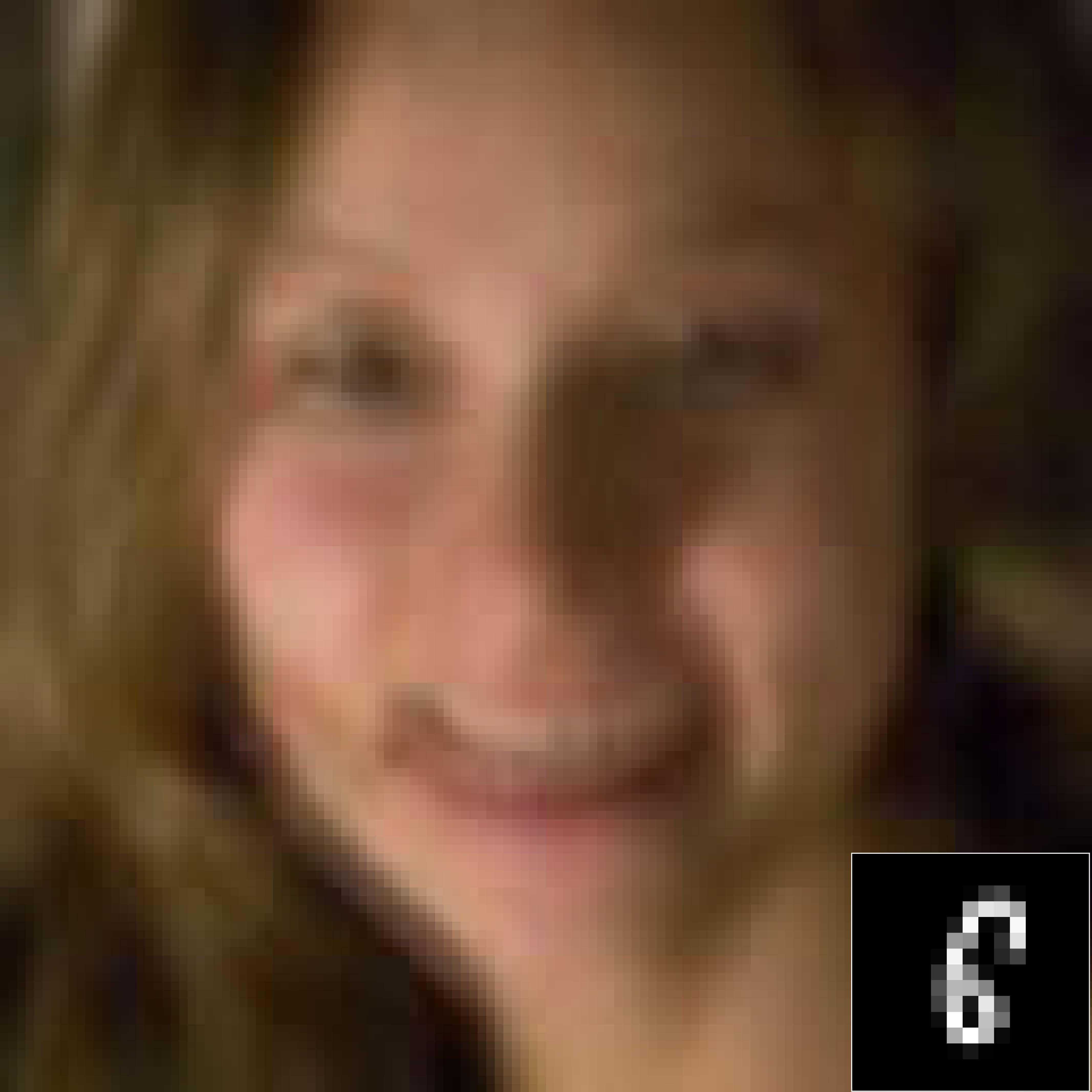} &
        \includegraphics[width=0.19\columnwidth]{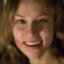} &
        \includegraphics[width=0.19\columnwidth]{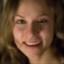} &
        \includegraphics[width=0.19\columnwidth]{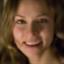}  
        \\
        \includegraphics[width=0.19\columnwidth]{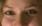} &
        \includegraphics[width=0.19\columnwidth]{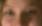} &
        \includegraphics[width=0.19\columnwidth]{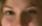} &
        \includegraphics[width=0.19\columnwidth]{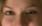} &
        \includegraphics[width=0.19\columnwidth]{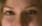}  
        \\
        \includegraphics[width=0.19\columnwidth]{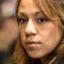} &
        \includegraphics[width=0.19\columnwidth]{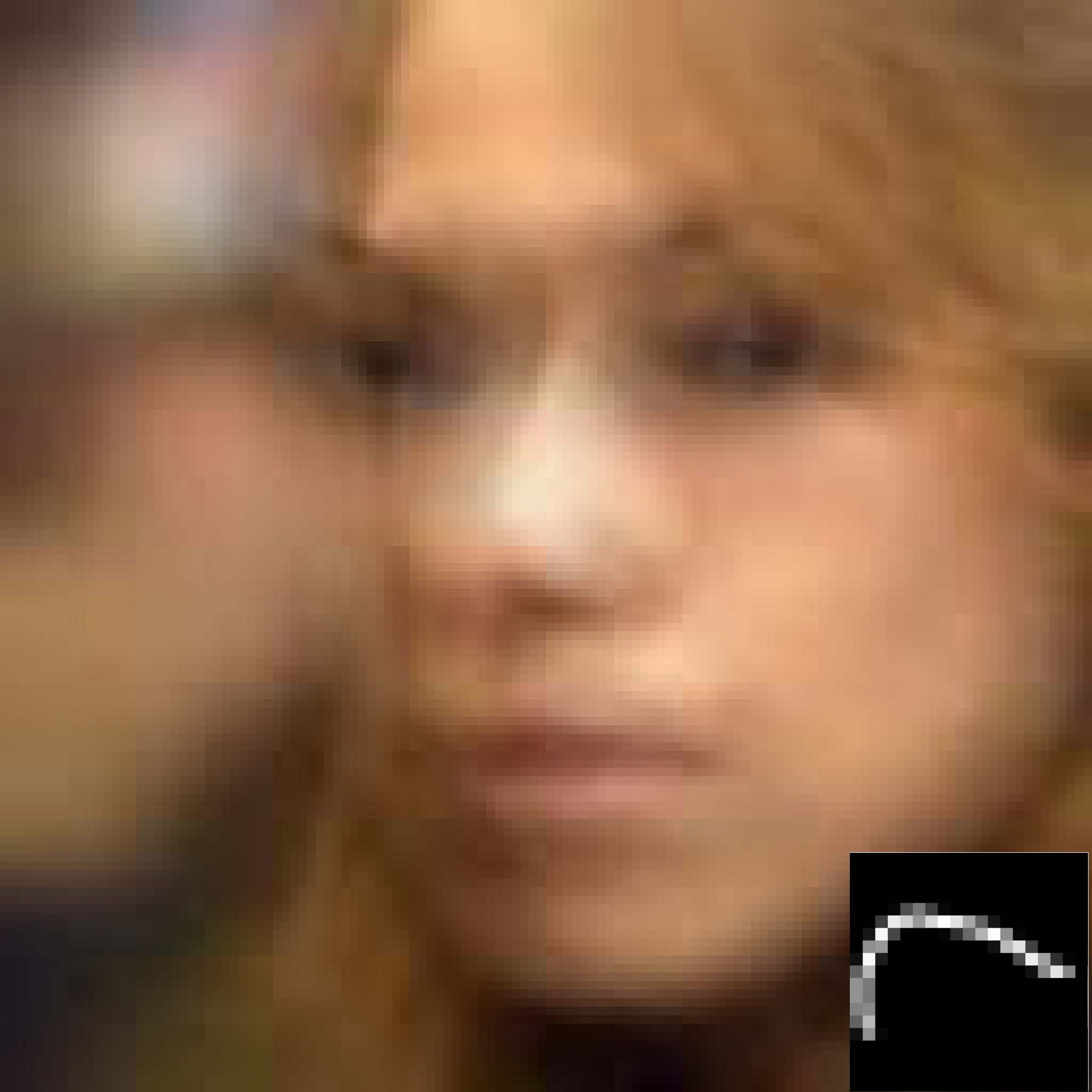} &
        \includegraphics[width=0.19\columnwidth]{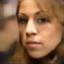} &
        \includegraphics[width=0.19\columnwidth]{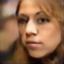} &
        \includegraphics[width=0.19\columnwidth]{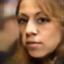}  
        \\
        \includegraphics[width=0.19\columnwidth]{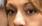} &
        \includegraphics[width=0.19\columnwidth]{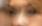} &
        \includegraphics[width=0.19\columnwidth]{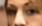} &
        \includegraphics[width=0.19\columnwidth]{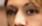} &
        \includegraphics[width=0.19\columnwidth]{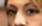}  
        \\
        (a) & (b) & (c) & (d) & (e) \\
    \end{tabular}
    \vspace{-2mm}
    \centering
    \caption{
        \textbf{Effects of loss functions.} 
        (a) Ground truth images
        (b) Blurred images
        (c) $\mathcal{L}_c$
        (d) $\mathcal{L}_c$ + equal-weight $\mathcal{L}_s$
        (e) $\mathcal{L}_c$ + adaptive $\mathcal{L}_s$
    }
    \label{fig:compare_loss}
    %\vspace{-2mm}
\end{figure}

\Paragraph{Local structural loss.}
\citet{Ziyi-CVPR-2018} adopt an \emph{equal} weight in the local structural loss $\mathcal{L}_s$ for all the key components, while we apply \emph{adaptive} weights based on the size of each component.
Here we train our fine deblurring network (freezing the coarse deblurring network and parsing network) using the content loss as well as local structural loss, and present the results in~\tabref{analysis_loss}.
We note that the model trained solely on the content loss $\mathcal{L}_c$ considers all the pixels, including hair, skin, and background, equally.
The equal-weight local structural loss significantly improves the performance by encouraging the network to enhance details on eight key components, including left eye, right eye, left eyebrow, right eyebrow, nose, upper lip, lower lip, and teeth.
The proposed adaptive local structural loss further adjusts the weights by considering the size of key components to prevent the model from sacrificing some tiny components, e.g., lips and teeth.
\figref{compare_loss} shows the deblurred results by the models listed in~\tabref{analysis_loss}.

\begin{table}[t]
	\centering
	\footnotesize
	\caption{
		\textbf{Analysis on loss functions}.
		We fix the parsing network and coarse deblurring network and train the fine deblurring network using the content loss $\mathcal{L}_c$ and local structural loss $\mathcal{L}_s$.
		}
	%\vspace{1mm}
	\begin{tabular}{l|cc|cc}
		\toprule
		\multirow{2}{*}{Losses} &
		\multicolumn{2}{c|}{Helen} &
		\multicolumn{2}{c}{CelebA} \\
		& PSNR & SSIM & PSNR & SSIM \\
		\midrule
        $\mathcal{L}_c$
        & 25.48  & 0.860 & 24.51 & 0.868 \\
		$\mathcal{L}_c$ + equal-weight $\mathcal{L}_s$
		& 25.72 & 0.863 & 24.72 & 0.869 \\
		$\mathcal{L}_c$ + adaptive $\mathcal{L}_s$
        & 25.80 & 0.866 & 24.86 & 0.874 \\
		\bottomrule
	\end{tabular}
	\label{tab:analysis_loss}
\end{table}

\begin{figure}
    \footnotesize
    \renewcommand{\tabcolsep}{1pt} % adjust horizontal space
    \renewcommand{\arraystretch}{0.5} % adjust vertical space
    \centering
    \begin{tabular}{ccccc}
        \includegraphics[width=0.19\columnwidth]{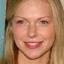} &
        \includegraphics[width=0.19\columnwidth]{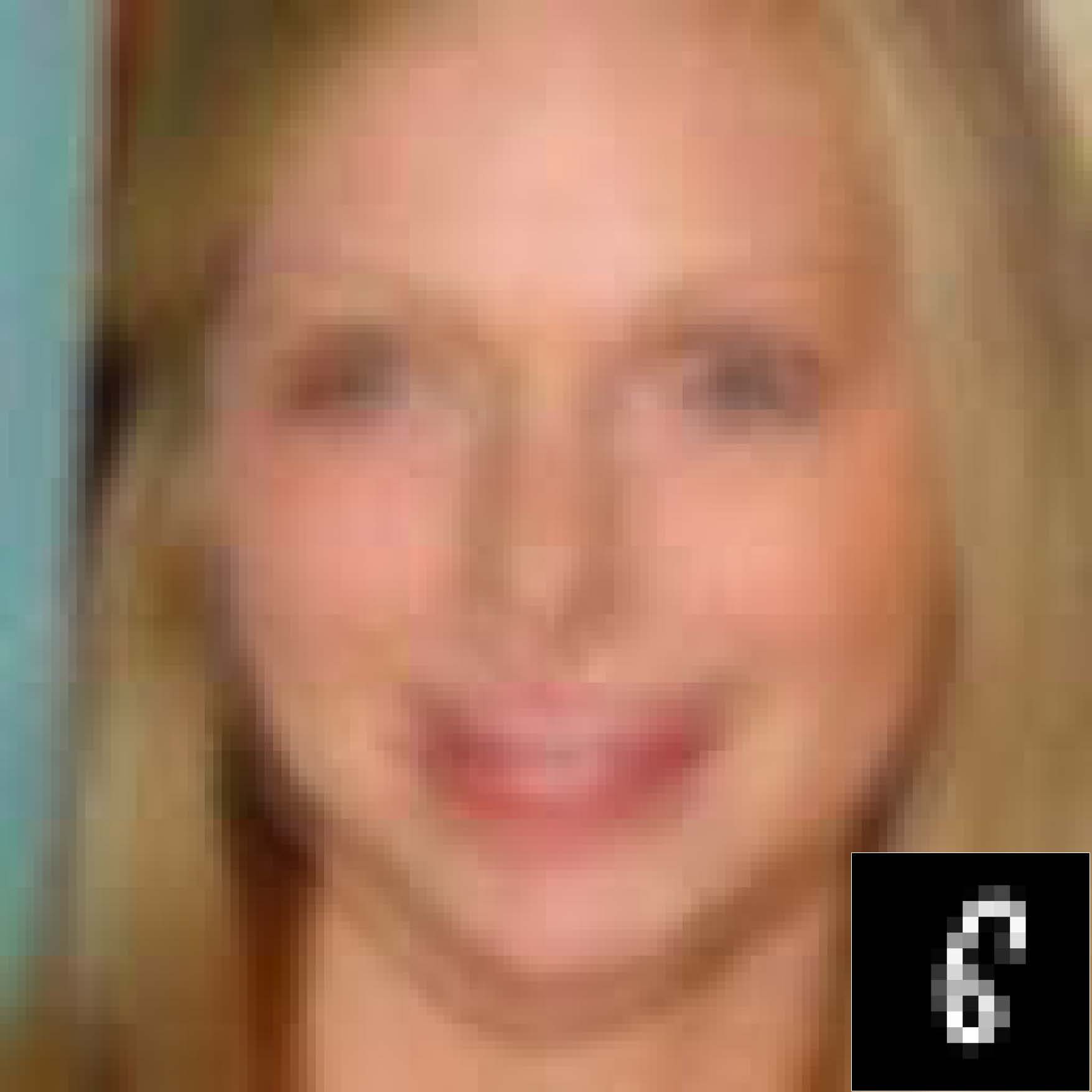} &
        \includegraphics[width=0.19\columnwidth]{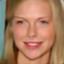} &
        \includegraphics[width=0.19\columnwidth]{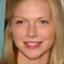} &
        \includegraphics[width=0.19\columnwidth]{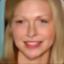}  
        \\
        \includegraphics[width=0.19\columnwidth]{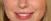} &
        \includegraphics[width=0.19\columnwidth]{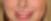} &
        \includegraphics[width=0.19\columnwidth]{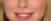} &
        \includegraphics[width=0.19\columnwidth]{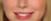} &
        \includegraphics[width=0.19\columnwidth]{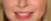}  
        \\
        \includegraphics[width=0.19\columnwidth]{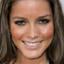} &
        \includegraphics[width=0.19\columnwidth]{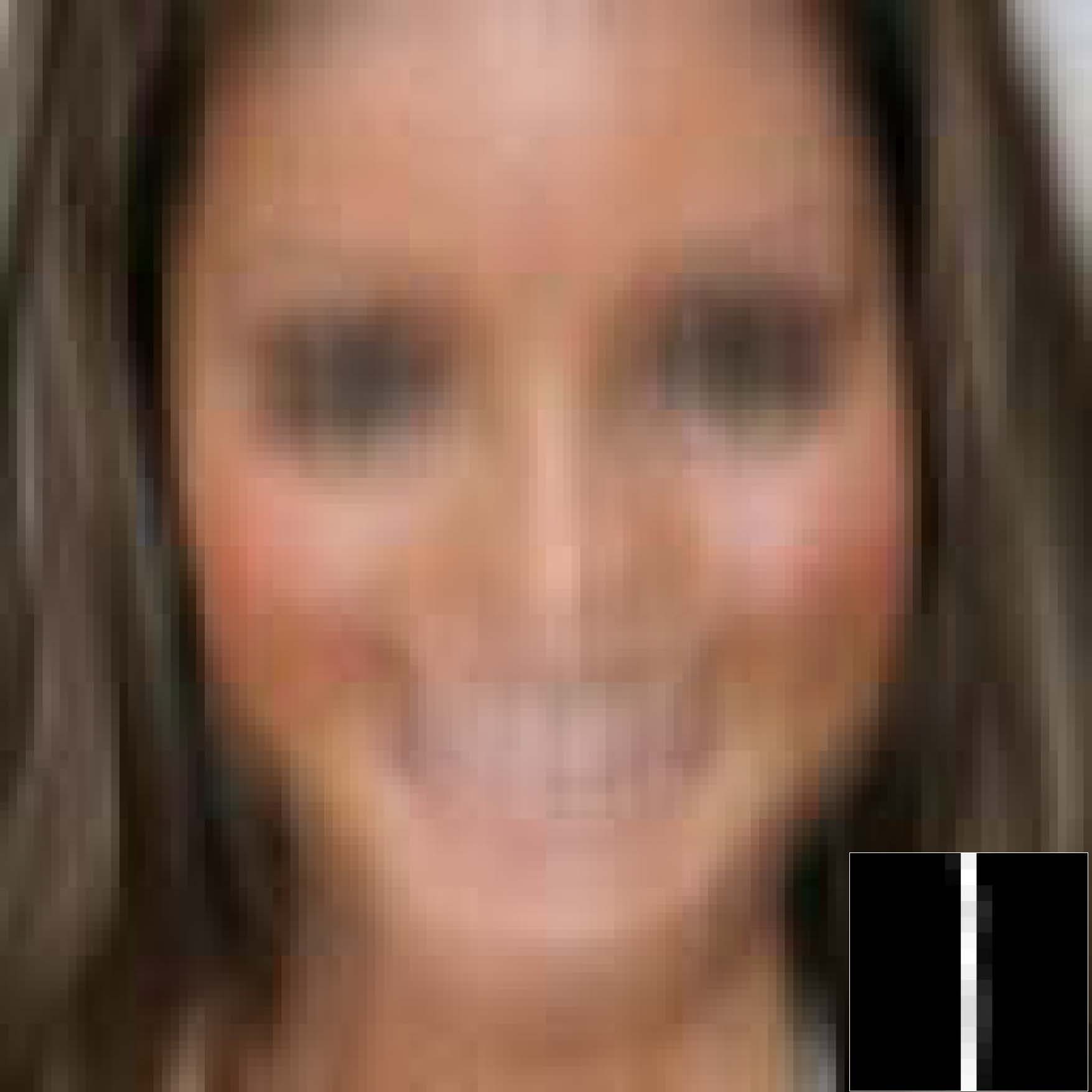} &
        \includegraphics[width=0.19\columnwidth]{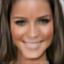} &
        \includegraphics[width=0.19\columnwidth]{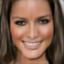} &
        \includegraphics[width=0.19\columnwidth]{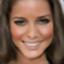}  
        \\
        \includegraphics[width=0.19\columnwidth]{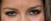} &
        \includegraphics[width=0.19\columnwidth]{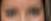} &
        \includegraphics[width=0.19\columnwidth]{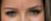} &
        \includegraphics[width=0.19\columnwidth]{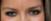} &
        \includegraphics[width=0.19\columnwidth]{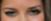}  
        \\
        (a) & (b) & (c) & (d) & (e) \\
    \end{tabular}
    \vspace{-2mm}
    \centering
    \caption{\textbf{Effects of training strategy.} 
        (a) Ground truth images
        (b) Blurred images
        (c) $\mathcal{G}_c$
        (d) $\mathcal{G}_c + \mathcal{P} + \mathcal{G}_f$ (fine-tuned)
        (e) $\mathcal{G}_c + \mathcal{P} + \mathcal{G}_f$ (from scratch)
    }
\label{fig:analysis_stage}
\end{figure}

\Paragraph{Training strategy.}
Since the proposed model consists of three sub-networks, the cascade of all sub-networks becomes a very deep model.
As such, it is not easy to training such a deep model from scratch.
The last row of~\tabref{analysis_stage} shows that the model trained from scratch does not perform well.
Thus, we train our model stage-by-stage using the training strategy described in~\secref{training_strategy}.
We show the evaluation results of each stage in~\tabref{analysis_stage}.
With the proposed training strategy, our model gradually achieves better performance.
\figref{analysis_stage} shows the deblurred results of the models listed in~\tabref{analysis_stage}.
The model using the progressive training strategy recovers better content and more facial details than the model trained from scratch.

\begin{table}[t]
	\centering
	\footnotesize
	\caption{
		\textbf{Analysis on training strategy}.
		We progressively train the coarse deblurring network $\mathcal{G}_c$, face parsing network $\mathcal{P}$, and the fine deblurring network $\mathcal{G}_f$.
		Finally, we jointly fine-tune all three sub-networks.
		The proposed training strategy achieves better performance than the training the whole model from scratch.
		}
	%\vspace{1mm}
	\begin{tabular}{c|cc|cc}
		\toprule
		\multirow{2}{*}{Model} &
		\multicolumn{2}{c|}{Helen} &
		\multicolumn{2}{c}{CelebA} \\
		& PSNR & SSIM & PSNR & SSIM \\
		\midrule
        $\mathcal{G}_c$
        & 25.26 & 0.855 & 24.58 & 0.869 \\
		$\mathcal{G}_c$ (fixed) + $\mathcal{P}$ (fixed) + $\mathcal{G}_f$
        & 25.80 & 0.866 & 24.86 & 0.874 \\
        $\mathcal{G}_c + \mathcal{P} + \mathcal{G}_f$ (fine-tuned)
		& 25.92 & 0.868 & 24.89 & 0.875 \\
		$\mathcal{G}_c + \mathcal{P} + \mathcal{G}_f$ (scratch)
		& 24.74 & 0.845 & 24.08 & 0.860 \\
		\bottomrule
	\end{tabular}
	\label{tab:analysis_stage}
\end{table}

\begin{figure}
	\footnotesize
	\renewcommand{\tabcolsep}{1pt} % adjust horizontal space
	\renewcommand{\arraystretch}{0.5} % adjust vertical space
	\centering
	\begin{tabular}{ccccc}
		\includegraphics[width=0.19\columnwidth]{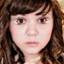} &
		\includegraphics[width=0.19\columnwidth]{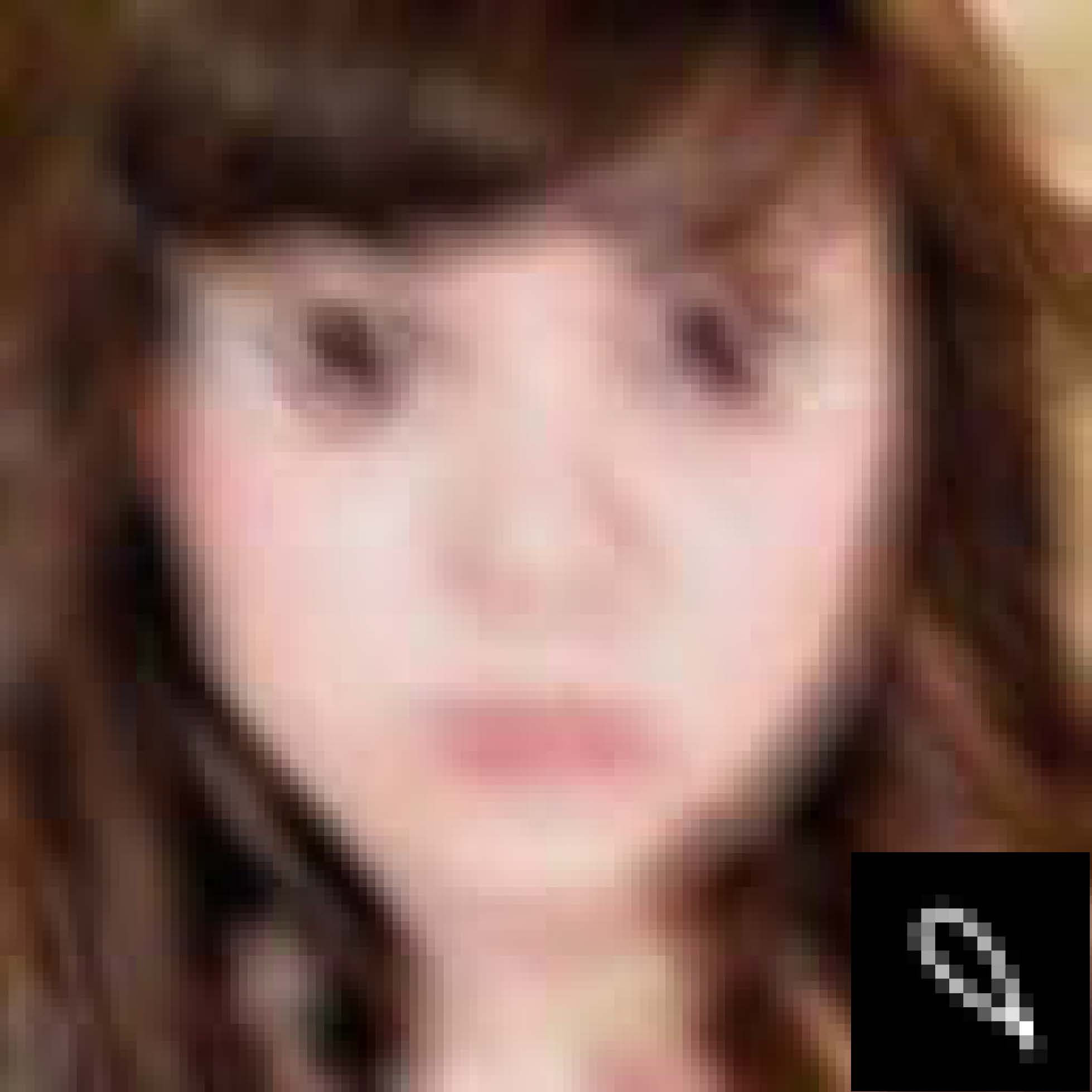} &
		\includegraphics[width=0.19\columnwidth]{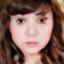} &
		\includegraphics[width=0.19\columnwidth]{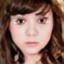} &
		\includegraphics[width=0.19\columnwidth]{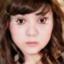}  
		\\
		\includegraphics[width=0.19\columnwidth]{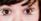} &
		\includegraphics[width=0.19\columnwidth]{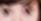} &
		\includegraphics[width=0.19\columnwidth]{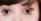} &
		\includegraphics[width=0.19\columnwidth]{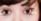} &
		\includegraphics[width=0.19\columnwidth]{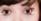}  
		\\
		\includegraphics[width=0.19\columnwidth]{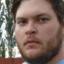} &
		\includegraphics[width=0.19\columnwidth]{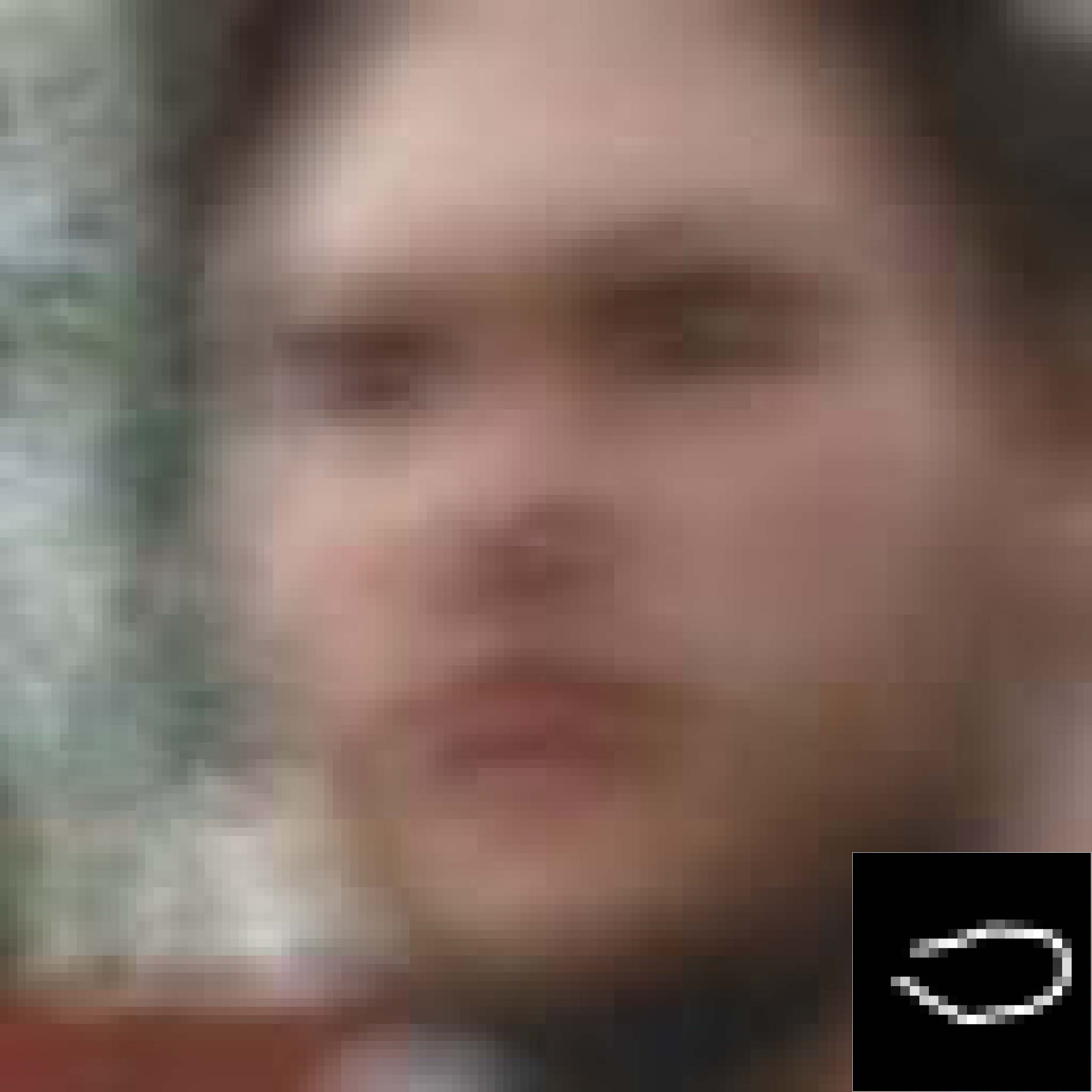} &
		\includegraphics[width=0.19\columnwidth]{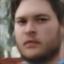} &
		\includegraphics[width=0.19\columnwidth]{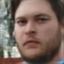} &
		\includegraphics[width=0.19\columnwidth]{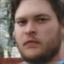}  
		\\
		\includegraphics[width=0.19\columnwidth]{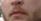} &
		\includegraphics[width=0.19\columnwidth]{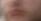} &
		\includegraphics[width=0.19\columnwidth]{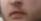} &
		\includegraphics[width=0.19\columnwidth]{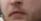} &
		\includegraphics[width=0.19\columnwidth]{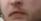}  
		\\
		(a) & (b) & (c) & (d) & (e)
	\end{tabular}
	\vspace{-1mm}
	\centering
	\caption{\textbf{Effects of perceptual and adversarial functions.} 
		(a) Ground truth images
		(b) Blurred images
		(c) Ours w/o $\mathcal{L}_{\text{VGG}}$ and $\mathcal{L}_{\text{adv}}$ 
		(d) Ours w/ $\mathcal{L}_{\text{VGG}}$ 
		(e) Ours w/ $\mathcal{L}_{\text{VGG}}$ and $\mathcal{L}_{\text{adv}}$ 
	}
	\label{fig:compare_vgg_adv}
	%\vspace{3mm}
\end{figure}

%%HERE
\Paragraph{Perceptual and adversarial losses.}
We compare the deblurring results with and without using the perceptual and adversarial losses in~\tabref{analysis_vgg_adv} and~\figref{compare_vgg_adv}.
The perceptual loss encourages the images to match the high-level activations of the VGG-Face network 
and makes the output look more photo-realistic.
The adversarial loss further introduces more details on hairs and beards, which cannot be reconstructed well using the pixel-wise $L_2$ or $L_1$ loss.
As shown in~\tabref{analysis_vgg_adv}, both the perceptual and adversarial losses improve the average PSNR and SSIM on both test sets as more faithful details are recovered.
\begin{table}[t]
	\centering
	\footnotesize
	\caption{
		\textbf{Analysis on perceptual and adversarial losses}.
		Both perceptual and adversarial losses further improve the performance by restoring more faithful details.
	}
	\begin{tabular}{cc|cc|cc}
		\toprule
		\multirow{2}{*}{$\mathcal{L}_{\text{VGG}}$} &
		\multirow{2}{*}{$\mathcal{L}_{\text{adv}}$} &
		\multicolumn{2}{c|}{Helen} & \multicolumn{2}{c}{CelebA} \\
		& & PSNR & SSIM & PSNR & SSIM \\
		\midrule
		& & 25.92 & 0.868 & 24.89 & 0.875 \\
		\checkmark & & 26.28 & 0.877 & 25.14 & 0.881 \\
		\checkmark & \checkmark & 26.34 & 0.876 & 25.33 & 0.881 \\
		\bottomrule
	\end{tabular}
	\label{tab:analysis_vgg_adv}
	\vspace{-2mm}
\end{table}

\Paragraph{Kernel size.}
We use a larger kernel size at the first convolutional layer of our coarse and fine deblurring networks.
Here we evaluate the performance of the proposed model with different kernel sizes in~\tabref{analysis_filter}.
Consistent performance gain can be achieved 
when using a larger filter size up to the kernel of $11 \times 11$ pixels.
Therefore, we choose to use $11 \times 11$ filter at the first convolutional layer to have a larger receptive field for the whole model.
In addition, as the first convolutional layer only contains 64 feature channels, using a larger filter size does not significantly increase the number of model parameters.

\begin{table}
	\centering
	\footnotesize
	\caption{
		\textbf{Analysis on kernel size}.
		We evaluate the model performance by changing the kernel size at the first convolutional layer.
	}
	\begin{tabular}{c|cc|cc|c}
		\toprule
		\multirow{2}{*}{Kernel size} &
		\multicolumn{2}{c|}{Helen} &
		\multicolumn{2}{c|}{CelebA} &
		\multirow{2}{*}{$\#$Parameters} \\
		& PSNR & SSIM & PSNR & SSIM & \\
		\midrule
		$5 \times 5$
		& 25.63 & 0.860 & 24.65 & 0.868 & 14.56 M \\
		$9 \times 9$
		& 25.75 & 0.864 & 24.72 & 0.868 & 14.70 M \\
		$11 \times 11$
		& 25.80 & 0.866 & 24.86 & 0.874 & 14.80 M \\
		$13 \times 13$
		&25.80  & 0.867 & 24.87 & 0.876 & 14.92 M\\
		\bottomrule
	\end{tabular}
	\label{tab:analysis_filter}
	\vspace{-2mm}
\end{table}

\begin{figure}
	\centering
	\begin{tabular}{cc}
		\includegraphics[width=0.47\columnwidth]{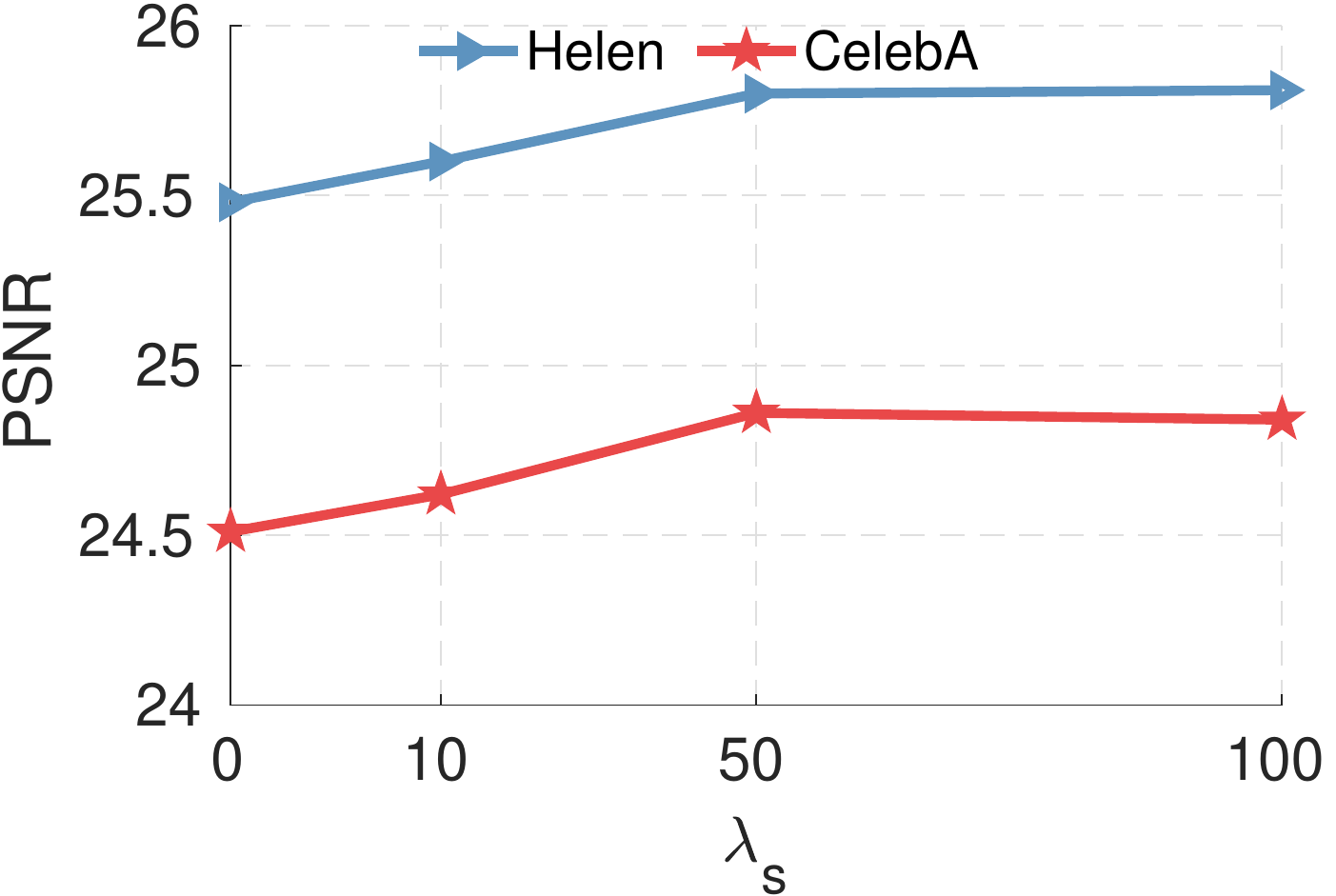} &
		\includegraphics[width=0.47\columnwidth]{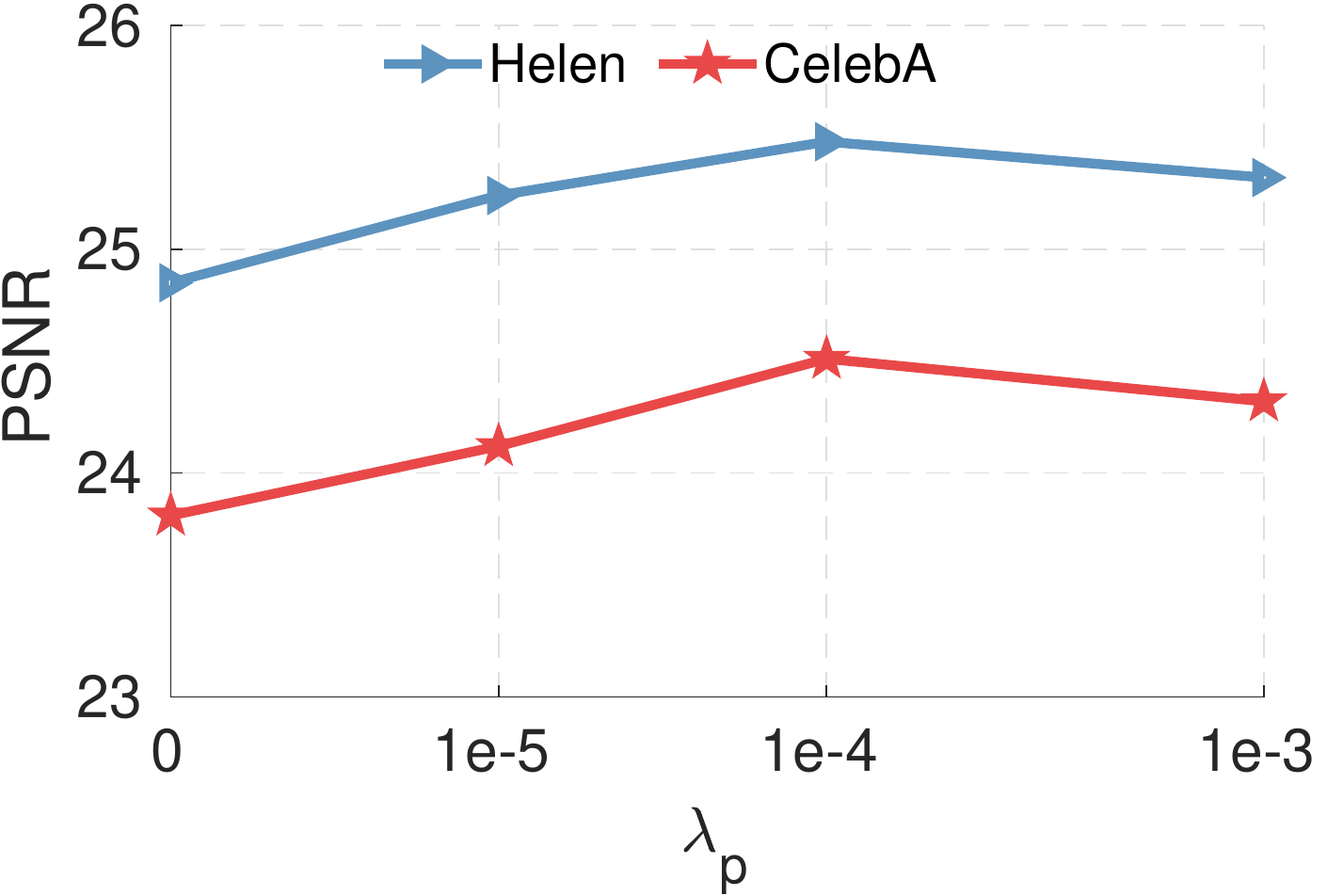} 
		\\
		(a) Local structural loss $\lambda_s$ &
		(b) Parsing loss $\lambda_p$
		\\
		\includegraphics[width=0.47\columnwidth]{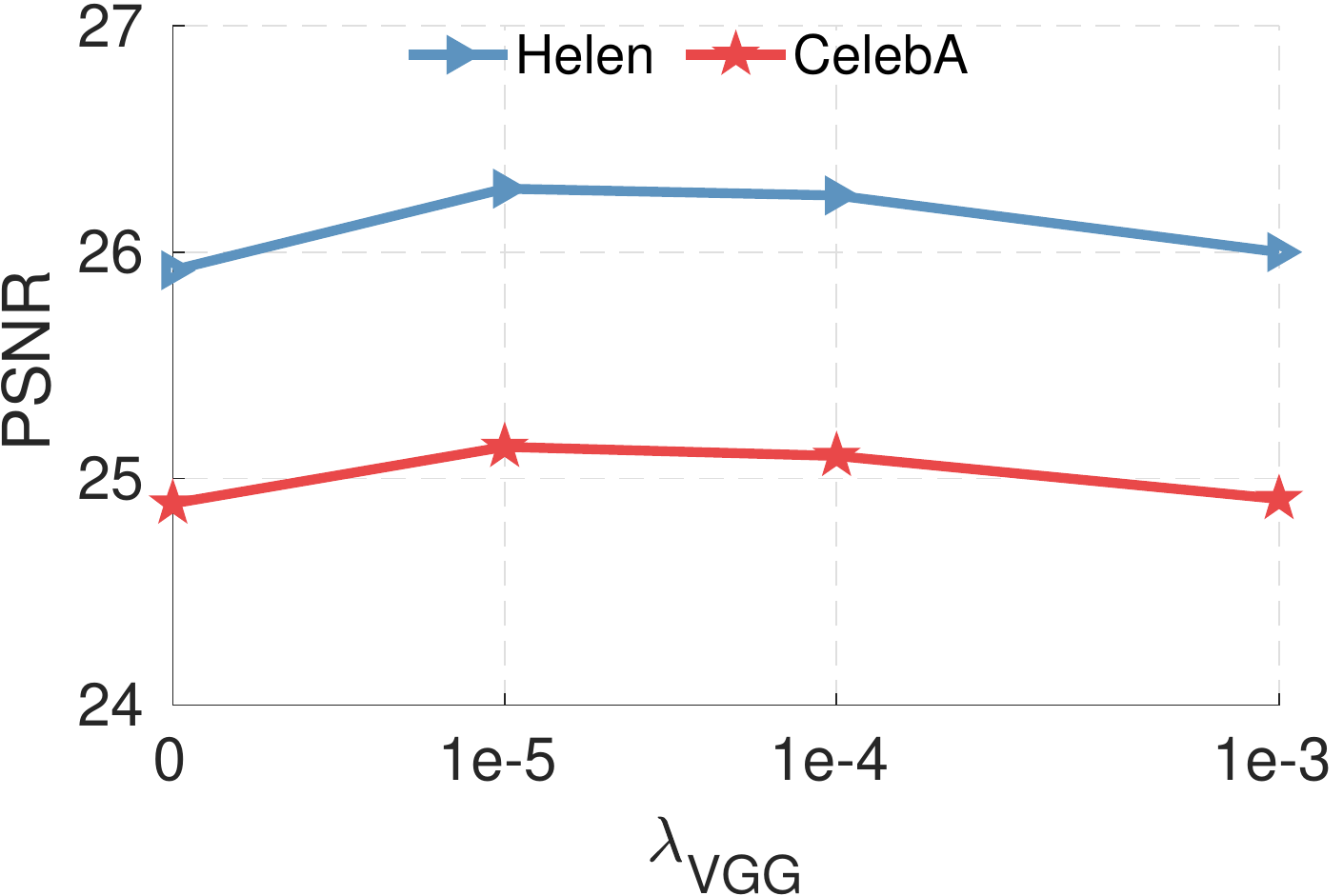} &
		\includegraphics[width=0.47\columnwidth]{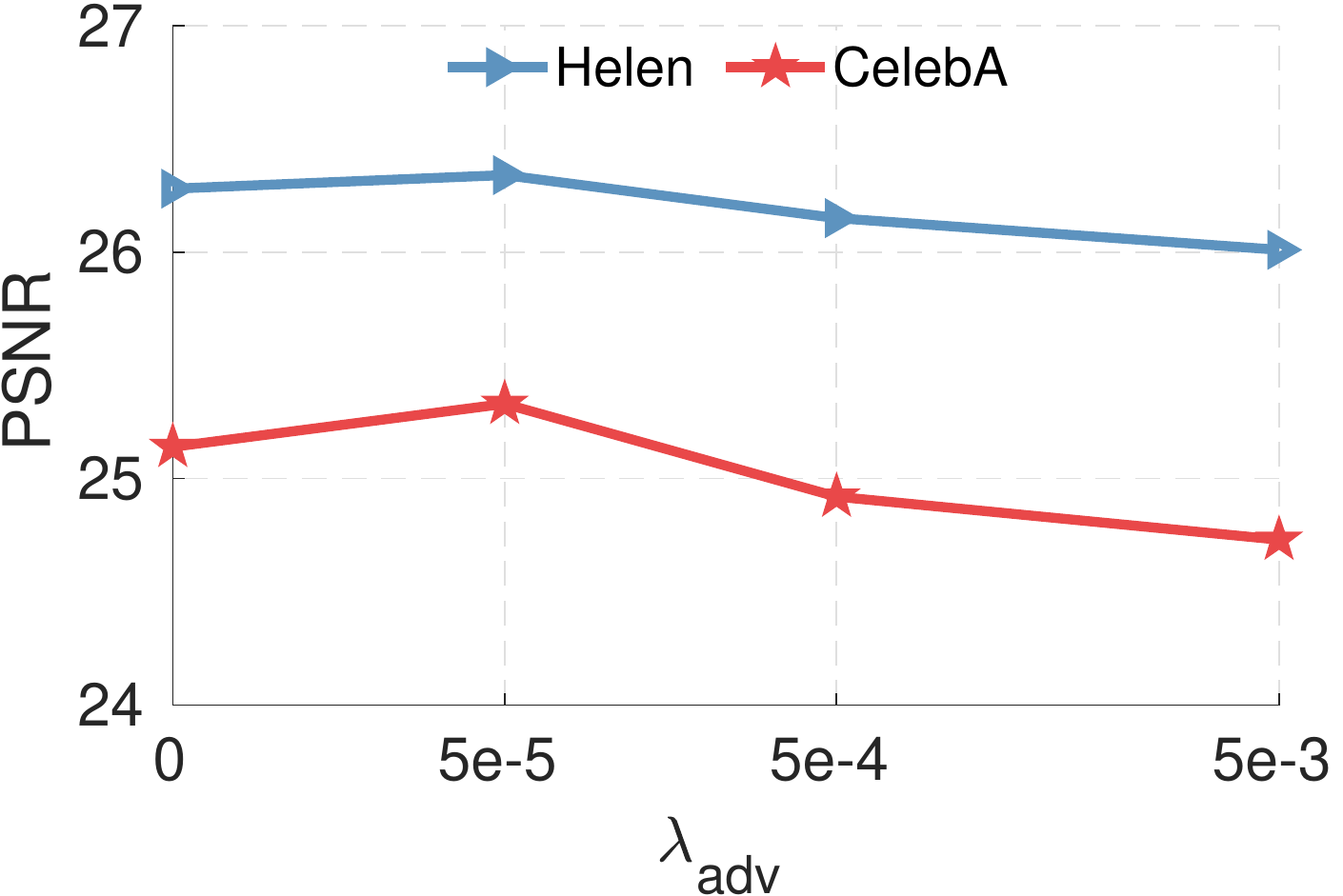} 
		\\
		(c) Perceptual loss $\lambda_{\text{VGG}}$ &
		(d) Adversarial loss $\lambda_{\text{adv}}$
	\end{tabular}
	\vspace{-1mm}
	\caption{
		\textbf{Effective range of hyper-parameters.}
		We plot the average PSNR on both the CelebA and Helen datasets.}
	\label{fig:weight}
	%\vspace{2mm}
\end{figure}

\Paragraph{Hyper-parameters.}
We analyze the effective range of the hyper-parameters, $\lambda_s$, $\lambda_p$, $\lambda_{\text{VGG}}$, and $\lambda_{\text{adv}}$ by changing one of the hyper-parameters and fixing the others.
In~\figref{weight}(a), we show that the local structural loss effectively improves the PSNR but saturates at $\lambda_s = 50$.
As shown in~\figref{weight}(b), without the parsing loss (i.e., $\lambda_p = 0$), the face parsing network cannot learn meaningful semantic labels as the facial priors.
However, a larger $\lambda_p$ does not further improve the face restoration performance as the gradient of the parsing loss is back-propagated to the coarse deblurring network, which may introduce additional artifacts.
Therefore, setting $\lambda_p = 1e-4$ achieves a good balance for the whole model.
In~\figref{weight}(c), we show that the proposed model obtains plausible results when choosing $1e-5 \leq \lambda_p \leq 1e-4$.
Using a larger weight for the perceptual loss introduces more checkerboard artifacts and harms the restoration performance.
Finally, in~\figref{weight}(d), we show that using a smaller weight for the adversarial loss, i.e., $\lambda_{\text{adv}} \leq 1e-4$, does not affect the PSNR too much.
However, the model can generate more facial details to improve visual quality.
When increasing $\lambda_{\text{adv}}$, the model generates noise-like artifacts, resulting in a performance drop.
Therefore, we choose $\lambda_{\text{adv}} = 5e-5$.

\Paragraph{Multi-stage deblurring.}

Due to our architecture design, we are able to extend the proposed model by cascading multiple fine deblurring networks.
Here we construct our model with one coarse deblurring network, one face parsing network, and $N$ fine deblurring networks.
The $N^{\text{th}}$ fine deblurring network takes as input the blurred image, deblurred image from the $N-1^{\text{th}}$ fine deblurring network, and the semantic labels from the face parsing network.
We compare the performance, model parameters, and execution time in  Table~\ref{tab:analysis_number}.
The performance of our model saturates at $N = 2$.
When using three fine deblurring networks, the model only slightly improves the performance but uses $160\%$ more parameters and runs $1.6\times$ slower than the model with $N = 1$.

In the last two rows of Table~\ref{tab:analysis_number}, we show the performance of the proposed model by sharing the weight of the fine deblurring network.
The experimental results show that the models with shared weights do not perform well.
As the fine deblurring network is already a deep sub-network, sharing the weight of a large sub-module is not guaranteed to improve the performance.
Instead, sharing the weight of a single convolutional layer or a small block (e.g., a residual block) might be a more reasonable way to design a recurrent structure.
As our goal is to utilize the semantic labels for face deblurring instead of exploring a better network architecture, we leave this issue as future work.
Overall, the proposed model with a single fine deblurring network already achieves state-of-the-art performance.

\begin{table}
	\centering
	\scriptsize
	\caption{
		\textbf{Multi-stage deblurring.}
		We apply the fine deblurring network for multiple times and compare the restoration performance, model parameters, and execution time.
	}
	%\vspace{-1mm}
	\begin{tabular}{c|cc|cc|cc}
		\toprule
		\multirow{2}{*}{$\#$Stages} &
		\multicolumn{2}{c|}{Helen} &
		\multicolumn{2}{c|}{CelebA} &
		\multirow{2}{*}{$\#$Parameters} &
		Time \\
		& PSNR & SSIM & PSNR & SSIM & & (s)\\
		\midrule
		$1$ & 25.80 & 0.866 & 24.86 & 0.874 & 14.80 M & 0.08 \\
		$2$ & 25.87 & 0.869 & 24.88 & 0.878 & 26.66 M & 0.11 \\
		$3$ & 25.89 & 0.866 & 24.86 & 0.875 & 38.52 M & 0.13 \\
		\midrule
		$2$ (shared) & 25.78 & 0.864 & 24.74 & 0.870 & 14.80 M & 0.11 \\
		$3$ (shared) & 25.74 & 0.861 & 24.76 & 0.871 & 14.80 M & 0.13 \\
		\bottomrule
	\end{tabular}
	\label{tab:analysis_number}
\end{table}

\begin{table*}
	\centering
	\scriptsize
	\caption{
		\textbf{Quantitative comparison with the state-of-the-art methods.}
		We compute the average PSNR and SSIM on the Helen and CelebA test sets.
		Each dataset has 8,000 blurred images synthesized from 100 clear face images and 80 blur kernels (10 blur kernels for each size).
		The \first{red} and \second{blue} texts indicate the best and second best performance.
	}
	\label{tab:evaluate_by_kernel}
	\resizebox{1.0\textwidth}{!}{
		%\vspace{1mm}
		\begin{tabular}{r|cccccccccccccccc}
			\toprule
			\multirow{2}{*}{Method} & 
			\multicolumn{2}{c}{$13 \times 13$} & 
			\multicolumn{2}{c}{$15 \times 15$} & 
			\multicolumn{2}{c}{$17 \times 17$} & 
			\multicolumn{2}{c}{$19 \times 19$} & 
			\multicolumn{2}{c}{$21 \times 21$} & 
			\multicolumn{2}{c}{$23 \times 23$} & 
			\multicolumn{2}{c}{$25 \times 25$} & 
			\multicolumn{2}{c}{$27 \times 27$}
			\\
			& PSNR & SSIM & PSNR & SSIM & PSNR & SSIM & PSNR & SSIM & PSNR & SSIM & PSNR & SSIM & PSNR & SSIM & PSNR & SSIM \\
			\midrule
			\multicolumn{17}{c}{Helen}\\
			\midrule
			\citet{DBLP:journals/tog/ShanJA08}
			& 20.65 & 0.726 & 20.01 & 0.693 & 20.66 & 0.718 & 19.16 & 0.653 & 19.75 & 0.675 & 19.07 & 0.645 & 19.51 & 0.665 & 17.79 & 0.587\\
			\citet{DBLP:journals/tog/ChoL09}
			& 17.46 & 0.610 & 17.16 & 0.590 & 17.58 & 0.613 & 16.58 & 0.563 & 16.82 & 0.574 & 16.52 & 0.554 & 16.88 & 0.575 & 15.58 & 0.509\\
			\citet{DBLP:conf/cvpr/KrishnanTF11}
			& 18.93 & 0.674 & 19.47 & 0.685 & 20.10 & 0.707 & 18.99 & 0.660 & 19.98 & 0.693 & 19.02 & 0.648 & 19.75 & 0.682 & 18.12 & 0.608\\
			\citet{DBLP:conf/cvpr/XuZJ13}
			& 20.85 & 0.744 & 20.51 & 0.730 & 21.12 & 0.751 & 19.43 & 0.686 & 20.38 & 0.723 & 19.65 & 0.691 & 20.24 & 0.714 & 18.69 & 0.648\\
			\citet{DBLP:conf/cvpr/ZhongCMPW13}
			& 17.33 & 0.646 & 16.83 & 0.631 & 16.78 & 0.629 & 16.38 & 0.613 & 15.96 & 0.598 & 16.07 & 0.600 & 16.02 & 0.603 & 15.98 & 0.593\\
			\citet{PAN17}
			& 23.00 & 0.797 & 22.01 & 0.769 & 22.87 & 0.792 & 20.35 & 0.711 & 20.99 & 0.733 & 19.90 & 0.691 & 20.32 & 0.708 & 18.04 & 0.617\\
			\citet{DBLP:conf/eccv/PanHSY14}
			&19.74&0.682&16.88&0.680&20.37&0.695&18.83&0.632&19.32&0.650&18.43&0.604&19.14&0.634&16.59&0.528\\
			\citet{Nah_2017_CVPR}
			& 26.02 & 0.878 & 25.43 & 0.858 & 25.84 & 0.862 & 23.37 & 0.805 & 24.45 & 0.835 & 23.10 & 0.793 & 23.80 & 0.815 & 20.98 & 0.734\\
			\citet{tao2018scale}
			&24.51&0.834&24.23&0.811&23.98&0.798&22.04&0.735&23.12&0.775&22.04&0.728&22.78&0.756&20.25&0.661\\
			\citet{li2018learning}
			&22.94&0.811&22.25&0.778&23.28&0.813&21.00&0.726&21.51&0.751&20.33&0.697&20.86&0.727&18.14&0.594\\
			\citet{deblurgan}
			&25.91&0.861&24.88&0.828&25.12&0.828&23.36&0.776&23.43&0.779&22.65&0.743&23.07&0.763&20.64&0.670\\
			\citet{Jin18}
			&26.75&0.897&26.09&0.880&\second{27.03}&\second{0.894}&24.64&0.847&25.38&0.863&\second{24.89}&\second{0.854}&\second{25.21}&\second{0.859}&\second{22.76}&\second{0.796}\\
			
			\citet{Ziyi-CVPR-2018}
			& \second{27.15} & \second{0.896} & \second{26.43} & \second{0.882} & 26.74 & 0.885 & \second{25.41} & \second{0.859} & \second{25.61} & \second{0.864} & 24.78 & 0.843 & 25.18 & 0.854 & 22.52 & 0.788\\
			Ours
			& \first{27.58} & \first{0.905} & \first{26.87} & \first{0.891} & \first{27.31} & \first{0.895} & \first{26.00} & \first{0.872} &
			\first{25.99} & \first{0.871} & \first{25.14} & \first{0.853} & \first{25.48} & \first{0.861} & \first{22.96} & \first{0.799}\\
			\midrule
			\multicolumn{17}{c}{CelebA}\\
			\midrule
			\citet{DBLP:journals/tog/ShanJA08}
			& 19.38 & 0.685 & 18.85 & 0.663 & 19.25 & 0.675 & 18.05 & 0.629 & 18.62 & 0.653 & 17.98 & 0.624 & 18.50 & 0.646 & 16.83 & 0.579\\
			\citet{DBLP:journals/tog/ChoL09}
			& 13.72 & 0.481 & 13.41 & 0.467 & 13.46 & 0.472 & 13.04 & 0.449 & 12.87 & 0.438 & 12.73 & 0.425 & 12.84 & 0.435 & 12.19 & 0.395\\
			\citet{DBLP:conf/cvpr/KrishnanTF11}
			& 18.23 & 0.679 & 18.57 & 0.685 & 19.18 & 0.705 & 18.04 & 0.661 & 19.18 & 0.699 & 17.98 & 0.648 & 18.77 & 0.684 & 17.12 & 0.617\\
			\citet{DBLP:conf/cvpr/XuZJ13}
			& 19.44 & 0.695 & 19.14 & 0.691 & 19.59 & 0.709 & 18.28 & 0.659 & 19.32 & 0.705 & 18.54 & 0.669 & 19.15 & 0.696 & 17.95 & 0.653\\
			\citet{DBLP:conf/cvpr/ZhongCMPW13}
			& 18.10 & 0.723 & 17.48 & 0.703 & 17.63 & 0.710 & 17.08 & 0.692 & 16.80 & 0.681 & 17.17 & 0.689 & 16.95 & 0.685 & 16.90 & 0.679\\
			\citet{PAN17}
			& 21.08 & 0.763 & 19.61 & 0.717 & 20.07 & 0.732 & 18.12 & 0.665 & 18.41 & 0.677 & 17.63 & 0.641 & 17.86 & 0.651 & 15.94 & 0.572\\
			\citet{DBLP:conf/eccv/PanHSY14}
			&20.47&0.712&20.60&0.709&21.16&0.730&19.36&0.655&19.89&0.677&18.82&0.628&19.51&0.664&16.78&0.540 \\
			\citet{Nah_2017_CVPR}
			& 24.23 & 0.879 & 23.49 & 0.862 & 23.87 & 0.863 & 21.90 & 0.820 & 22.65 & 0.839 & 21.53 & 0.806 & 22.12 & 0.824 & 19.71 & 0.759\\
			\citet{tao2018scale}
			&26.10&0.881&25.49&0.853&25.7&0.856&23.13&0.771&24.54&0.826&23.05&0.768&23.89&0.801&21.04&0.694\\
			\citet{li2018learning}
			&20.36&0.724&18.92&0.664&18.68&0.663&17.19&0.589&16.79&0.575&16.35&0.541&16.49&0.556&14.90&0.461\\
			\citet{deblurgan}
			&24.34&0.802&23.36&0.765&23.54&0.764&22.31&0.726&22.31&0.727&21.68&0.696&22.06&0.715&20.01&0.639\\
			\citet{Jin18}
			&26.08&0.897&25.50&0.843&\second{26.01}&0.880&24.02&0.821&\second{24.79}& \second{0.871}&23.79&0.831&24.10&0.836&21.42&0.747\\
			\citet{Ziyi-CVPR-2018}
			& \second{26.19} & \second{0.900} & \second{25.52} & \second{0.887} & 25.77 & \second{0.889} & \second{24.58} & \second{0.867} & 24.61 & 0.868 & \second{23.87} & \second{0.852} & \second{24.25} & \second{0.861} & \second{21.93} & \second{0.807}\\
			Ours
			& \first{26.25} & \first{0.905} & \first{25.71} & \first{0.894} & \first{26.14} & \first{0.898} & \first{25.01} & \first{0.878} &
			\first{25.04} & \first{0.879} & \first{24.24} & \first{0.862} & \first{24.56} & \first{0.870} & \first{22.19} & \first{0.814}\\
			\bottomrule
		\end{tabular}}
		%\vspace{2mm}
	\end{table*}

\begin{table*}
	\centering
	\scriptsize
	\caption{
		\textbf{Quantitative comparison with the state-of-the-art methods.}
		We compute the average PSNR and SSIM on the Helen and CelebA test sets.
		The \first{red} and \second{blue} texts indicate the best and second best performance.
	}
	\label{tab:average_psnr_ssim}
	\resizebox{1.0\textwidth}{!}{
		\begin{tabular}{r|cccc|cccc}
			\toprule
			\multirow{2}{*}{Method} & 
			\multicolumn{4}{c|}{Helen} & 
			\multicolumn{4}{c}{CelebA} \\
			& Average PSNR & Worst PSNR & Average SSIM & Worst SSIM 
			& Average PSNR & Worst PSNR & Average SSIM & Worst SSIM
			\\
			\midrule
			\citet{DBLP:journals/tog/ShanJA08}
			& 19.57$\,\pm\,$2.72&9.97&0.670$\,\pm\,$0.137 & 0.109& 18.43$\,\pm\,$2.20 &  9.72&0.644$\,\pm\,$0.119&0.040\\
			\citet{DBLP:journals/tog/ChoL09}
			& 16.82$\,\pm\,$2.79& 7.83&0.574$\,\pm\,$0.126 & 0.215& 13.03$\,\pm\,$1.74 & 8.21&0.445$\,\pm\,$0.098&  0.097 \\
			\citet{DBLP:conf/cvpr/KrishnanTF11}
			& 19.30$\,\pm\,$3.42&  5.91&0.670$\,\pm\,$0.167 & 0.137& 18.38$\,\pm\,$2.74& 6.90& 0.672$\,\pm\,$0.146&0.108\\
			\citet{DBLP:conf/cvpr/XuZJ13}
			& 20.11$\,\pm\,$3.18& 9.45& 0.711$\,\pm\,$0.147& 0.075& 18.93$\,\pm\,$2.55& 9.92& 0.685$\,\pm\,$0.124&0.106\\
			\citet{DBLP:conf/cvpr/ZhongCMPW13}
			& 16.41$\,\pm\,$3.13& 8.25& 0.614$\,\pm\,$0.142& 0.140& 17.26$\,\pm\,$2.66& 10.41&0.695$\,\pm\,$0.115&0.278\\
			\citet{DBLP:conf/eccv/PanHSY14}
			& 18.66$\,\pm\,$3.95&8.82&0.677$\,\pm\,$0.175& 0.167&18.59$\,\pm\,$3.59&9.47&0.677$\,\pm\,$0.183&0.117\\
			\citet{PAN17}
			& 20.93$\,\pm\,$4.27& 7.21&0.727$\,\pm\,$0.168& 0.120& 19.57$\,\pm\,$4.02&8.07& 0.664$\,\pm\,$0.160&0.118\\
			\citet{Nah_2017_CVPR}
			& 24.12$\,\pm\,$3.46& 11.59& 0.823$\,\pm\,$0.107& 0.240& 22.43$\,\pm\,$2.82& 12.11& 0.832$\,\pm\,$0.103&0.277\\
			\citet{tao2018scale}
			& 22.86$\,\pm\,$3.51& 11.68& 0.762$\,\pm\,$0.109& 0.259& 24.11$\,\pm\,$2.67& 11.21&\second{0.862$\,\pm\,$0.091}&0.245\\
			\citet{li2018learning}
			& 21.28$\,\pm\,$3.65& 9.24& 0.737$\,\pm\,$0.143& 0.159&17.46$\,\pm\,$3.39& 8.86& 0.596$\,\pm\,$0.166& 0.129\\
			\citet{deblurgan}
			& 23.63$\,\pm\,$2.90& 11.98& 0.781$\,\pm\,$0.094 & 0.267& 22.45$\,\pm\,$2.21& 12.29& 0.729$\,\pm\,$0.080&0.283\\
			\citet{Jin18}
			& 25.34$\,\pm\,$3.17& 11.49&\second{0.861$\,\pm\,$0.078}& 0.252& \second{24.46$\,\pm\,$2.77}& 12.54& 0.841$\,\pm\,$0.075&0.308\\
			\citet{Ziyi-CVPR-2018}
			& \second{25.58$\,\pm\,$2.94}& 12.45& \second{0.861$\,\pm\,$0.070}& 0.403& 24.34$\,\pm\,$2.46& 12.03&0.860$\,\pm\,$0.066&0.303\\
			Ours
			& \first{25.91$\,\pm\,$2.91}& 13.55& \first{0.869$\,\pm\,$0.062}& 0.480& \first{24.89$\,\pm\,$2.32}& 12.46& \first{0.875$\,\pm\,$0.063}&0.310\\
			\bottomrule
		\end{tabular}
	}
\end{table*}

\begin{figure*}
	\scriptsize
	\renewcommand{\tabcolsep}{1pt} % adjust horizontal space
	\renewcommand{\arraystretch}{0.8} % adjust vertical space
	\centering
	\begin{tabular}{ccccccccccc}
		\includegraphics[width=0.087\textwidth]{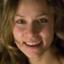}  &
		\includegraphics[width=0.087\textwidth]{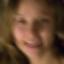} &
		\includegraphics[width=0.087\textwidth]{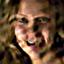} &
		\includegraphics[width=0.087\textwidth]{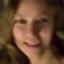} &
		\includegraphics[width=0.087\textwidth]{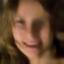} &
		\includegraphics[width=0.087\textwidth]{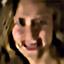} &
		\includegraphics[width=0.087\textwidth]{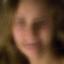} &
		\includegraphics[width=0.087\textwidth]{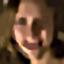} &
		\includegraphics[width=0.087\textwidth]{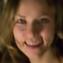} &
		\includegraphics[width=0.087\textwidth]{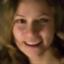} &
		\includegraphics[width=0.087\textwidth]{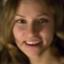} 
		\\
		\includegraphics[width=0.087\textwidth]{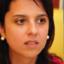} &
		\includegraphics[width=0.087\textwidth]{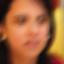} &
		\includegraphics[width=0.087\textwidth]{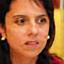} &
		\includegraphics[width=0.087\textwidth]{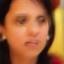} &
		\includegraphics[width=0.087\textwidth]{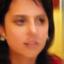} &
		\includegraphics[width=0.087\textwidth]{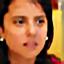} &
		\includegraphics[width=0.087\textwidth]{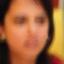} &
		\includegraphics[width=0.087\textwidth]{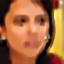} &
		\includegraphics[width=0.087\textwidth]{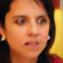} &
		\includegraphics[width=0.087\textwidth]{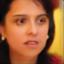} &
		\includegraphics[width=0.087\textwidth]{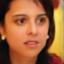}  
		\\
		\includegraphics[width=0.087\textwidth]{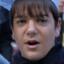} &
		\includegraphics[width=0.087\textwidth]{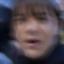} &
		\includegraphics[width=0.087\textwidth]{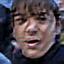} &
		\includegraphics[width=0.087\textwidth]{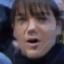} &
		\includegraphics[width=0.087\textwidth]{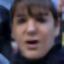} &
		\includegraphics[width=0.087\textwidth]{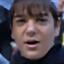} &
		\includegraphics[width=0.087\textwidth]{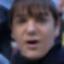} &
		\includegraphics[width=0.087\textwidth]{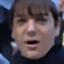} &
		\includegraphics[width=0.087\textwidth]{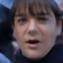} &
		\includegraphics[width=0.087\textwidth]{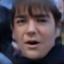} &
		\includegraphics[width=0.087\textwidth]{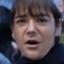} 
		\\
		\includegraphics[width=0.087\textwidth]{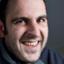} &
		\includegraphics[width=0.087\textwidth]{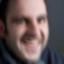} &
		\includegraphics[width=0.087\textwidth]{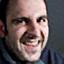} &
		\includegraphics[width=0.087\textwidth]{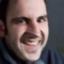} &
		\includegraphics[width=0.087\textwidth]{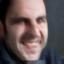} &
		\includegraphics[width=0.087\textwidth]{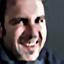} &
		\includegraphics[width=0.087\textwidth]{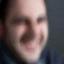} &
		\includegraphics[width=0.087\textwidth]{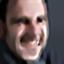} &
		\includegraphics[width=0.087\textwidth]{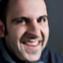} &
		\includegraphics[width=0.087\textwidth]{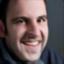} &
		\includegraphics[width=0.087\textwidth]{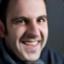} 
		\\
		(a) GT &
		(b) Blurred &
		(c) \citeauthor{DBLP:conf/eccv/PanHSY14} &
		(d) \citeauthor{Nah_2017_CVPR} &
		(e) \citeauthor{tao2018scale} &
		(f) \citeauthor{li2018learning} &
		(g) \citeauthor{deblurgan} &
		(h) \citeauthor{PAN17} &
		(i) \citeauthor{Jin18}&
		(j) \citeauthor{Ziyi-CVPR-2018} &
		(k) Ours 
		\\
		& 
		& 
		\citeyear{DBLP:conf/eccv/PanHSY14} &
		\citeyear{Nah_2017_CVPR} &
		\citeyear{tao2018scale} &
		\citeyear{li2018learning} &
		\citeyear{deblurgan} &
		\citeyear{PAN17} &
		\citeyear{Jin18}&
		\citeyear{Ziyi-CVPR-2018} &
		\\
	\end{tabular}
	\vspace{-1mm}
	\caption{
		\textbf{Visual comparison on Helen dataset.}
		The results from the proposed method contain fewer visual artifacts and more details on key face components (e.g., eyes and mouths).
	}
	\label{fig:visual_comparison_helen}
\end{figure*}

\begin{figure*}
	\scriptsize
	\renewcommand{\tabcolsep}{1pt} % adjust horizontal space
	\renewcommand{\arraystretch}{0.8} % adjust vertical space
	\centering
	\begin{tabular}{ccccccccccc}
		\includegraphics[width=0.087\textwidth]{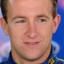}  &
		\includegraphics[width=0.087\textwidth]{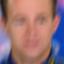} &
		\includegraphics[width=0.087\textwidth]{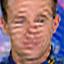} &
		\includegraphics[width=0.087\textwidth]{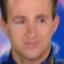} &
		\includegraphics[width=0.087\textwidth]{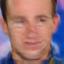} &
		\includegraphics[width=0.087\textwidth]{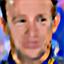} &
		\includegraphics[width=0.087\textwidth]{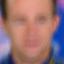} &
		\includegraphics[width=0.087\textwidth]{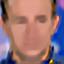} &
		\includegraphics[width=0.087\textwidth]{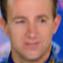} &
		\includegraphics[width=0.087\textwidth]{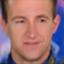} &
		\includegraphics[width=0.087\textwidth]{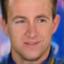} 
		\\
		\includegraphics[width=0.087\textwidth]{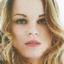} &
		\includegraphics[width=0.087\textwidth]{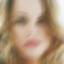} &
		\includegraphics[width=0.087\textwidth]{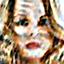} &
		\includegraphics[width=0.087\textwidth]{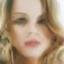} &
		\includegraphics[width=0.087\textwidth]{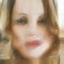} &
		\includegraphics[width=0.087\textwidth]{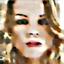} &
		\includegraphics[width=0.087\textwidth]{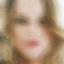} &
		\includegraphics[width=0.087\textwidth]{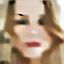} &
		\includegraphics[width=0.087\textwidth]{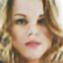} &
		\includegraphics[width=0.087\textwidth]{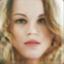} &
		\includegraphics[width=0.087\textwidth]{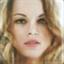}  
		\\
		\includegraphics[width=0.087\textwidth]{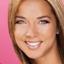} &
		\includegraphics[width=0.087\textwidth]{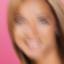} &
		\includegraphics[width=0.087\textwidth]{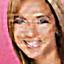} &
		\includegraphics[width=0.087\textwidth]{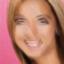} &
		\includegraphics[width=0.087\textwidth]{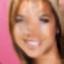} &
		\includegraphics[width=0.087\textwidth]{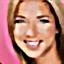} &
		\includegraphics[width=0.087\textwidth]{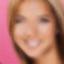} &
		\includegraphics[width=0.087\textwidth]{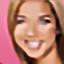} &
		\includegraphics[width=0.087\textwidth]{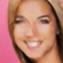} &
		\includegraphics[width=0.087\textwidth]{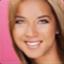} &
		\includegraphics[width=0.087\textwidth]{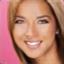} 
		\\
		\includegraphics[width=0.087\textwidth]{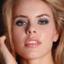} &
		\includegraphics[width=0.087\textwidth]{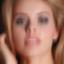} &
		\includegraphics[width=0.087\textwidth]{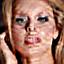} &
		\includegraphics[width=0.087\textwidth]{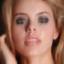} &
		\includegraphics[width=0.087\textwidth]{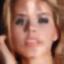} &
		\includegraphics[width=0.087\textwidth]{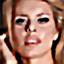} &
		\includegraphics[width=0.087\textwidth]{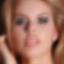} &
		\includegraphics[width=0.087\textwidth]{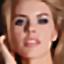} &
		\includegraphics[width=0.087\textwidth]{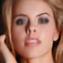} &
		\includegraphics[width=0.087\textwidth]{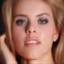} &
		\includegraphics[width=0.087\textwidth]{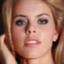} 
		\\
		(a) GT &
		(b) Blurred &
		(c) \citeauthor{DBLP:conf/eccv/PanHSY14} &
		(d) \citeauthor{Nah_2017_CVPR} &
		(e) \citeauthor{tao2018scale} &
		(f) \citeauthor{li2018learning} &
		(g) \citeauthor{deblurgan} &
		(h) \citeauthor{PAN17} &
		(i) \citeauthor{Jin18}&
		(j) \citeauthor{Ziyi-CVPR-2018} &
		(k) Ours 
		\\
		& 
		& 
		\citeyear{DBLP:conf/eccv/PanHSY14} &
		\citeyear{Nah_2017_CVPR} &
		\citeyear{tao2018scale} &
		\citeyear{li2018learning} &
		\citeyear{deblurgan} &
		\citeyear{PAN17} &
		\citeyear{Jin18} &
		\citeyear{Ziyi-CVPR-2018} &
		
		\\
	\end{tabular}
	\vspace{-1mm}
	\caption{
		\textbf{Visual comparison on CelebA dataset.}
		The results from the proposed method contain fewer visual artifacts and more details on key face components (e.g., eyes and mouths).
	}
	\label{fig:visual_comparison_celeba}
\end{figure*}

\begin{figure*}
	\scriptsize
	\renewcommand{\tabcolsep}{1pt} % adjust horizontal space
	\renewcommand{\arraystretch}{0.8} % adjust vertical space
	\centering
	\begin{tabular}{ccccccccccc}
		\includegraphics[width=0.087\textwidth]{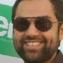} &
		\includegraphics[width=0.087\textwidth]{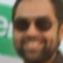} &
		\includegraphics[width=0.087\textwidth]{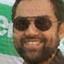} &
		\includegraphics[width=0.087\textwidth]{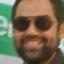} &
		\includegraphics[width=0.087\textwidth]{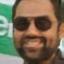} &
		\includegraphics[width=0.087\textwidth]{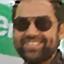} &
		\includegraphics[width=0.087\textwidth]{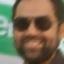} &
		\includegraphics[width=0.087\textwidth]{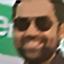} &
		\includegraphics[width=0.087\textwidth]{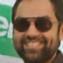} &
		\includegraphics[width=0.087\textwidth]{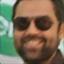} &
		\includegraphics[width=0.087\textwidth]{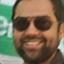}  
		\\
		\includegraphics[width=0.087\textwidth]{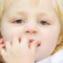} &
		\includegraphics[width=0.087\textwidth]{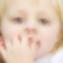} &
		\includegraphics[width=0.087\textwidth]{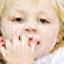} &
		\includegraphics[width=0.087\textwidth]{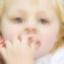} &
		\includegraphics[width=0.087\textwidth]{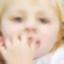} &
		\includegraphics[width=0.087\textwidth]{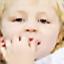} &
		\includegraphics[width=0.087\textwidth]{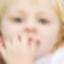} &
		\includegraphics[width=0.087\textwidth]{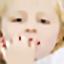} &
		\includegraphics[width=0.087\textwidth]{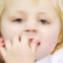} &
		\includegraphics[width=0.087\textwidth]{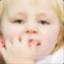} &
		\includegraphics[width=0.087\textwidth]{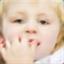} 
		\\
		\includegraphics[width=0.087\textwidth]{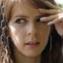} &
		\includegraphics[width=0.087\textwidth]{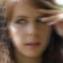} &
		\includegraphics[width=0.087\textwidth]{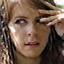} &
		\includegraphics[width=0.087\textwidth]{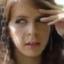} &
		\includegraphics[width=0.087\textwidth]{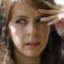} &
		\includegraphics[width=0.087\textwidth]{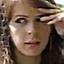} &
		\includegraphics[width=0.087\textwidth]{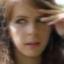} &
		\includegraphics[width=0.087\textwidth]{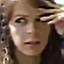} &
		\includegraphics[width=0.087\textwidth]{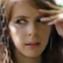} &
		\includegraphics[width=0.087\textwidth]{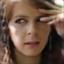} &
		\includegraphics[width=0.087\textwidth]{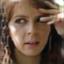} 
		\\
		\includegraphics[width=0.087\textwidth]{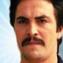}  &
		\includegraphics[width=0.087\textwidth]{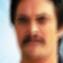} &
		\includegraphics[width=0.087\textwidth]{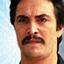} &
		\includegraphics[width=0.087\textwidth]{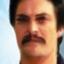} &
		\includegraphics[width=0.087\textwidth]{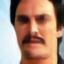} &
		\includegraphics[width=0.087\textwidth]{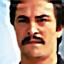} &
		\includegraphics[width=0.087\textwidth]{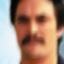} &
		\includegraphics[width=0.087\textwidth]{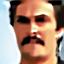} &
		\includegraphics[width=0.087\textwidth]{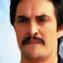} &
		\includegraphics[width=0.087\textwidth]{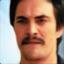} &
		\includegraphics[width=0.087\textwidth]{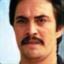} 
		\\
		\includegraphics[width=0.087\textwidth]{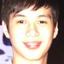} &
		\includegraphics[width=0.087\textwidth]{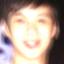} &
		\includegraphics[width=0.087\textwidth]{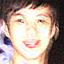} &
		\includegraphics[width=0.087\textwidth]{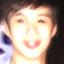} &
		\includegraphics[width=0.087\textwidth]{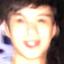} &
		\includegraphics[width=0.087\textwidth]{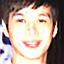} &
		\includegraphics[width=0.087\textwidth]{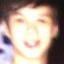} &
		\includegraphics[width=0.087\textwidth]{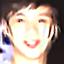} &
		\includegraphics[width=0.087\textwidth]{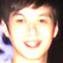} &
		\includegraphics[width=0.087\textwidth]{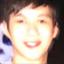} &
		\includegraphics[width=0.087\textwidth]{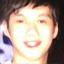} 
		\\
		\includegraphics[width=0.087\textwidth]{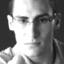} &
		\includegraphics[width=0.087\textwidth]{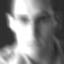} &
		\includegraphics[width=0.087\textwidth]{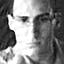} &
		\includegraphics[width=0.087\textwidth]{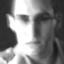} &
		\includegraphics[width=0.087\textwidth]{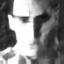} &
		\includegraphics[width=0.087\textwidth]{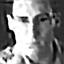} &
		\includegraphics[width=0.087\textwidth]{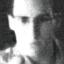} &
		\includegraphics[width=0.087\textwidth]{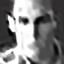} &
		\includegraphics[width=0.087\textwidth]{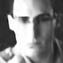} &
		\includegraphics[width=0.087\textwidth]{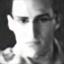} &
		\includegraphics[width=0.087\textwidth]{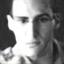} 
		\\
		\includegraphics[width=0.087\textwidth]{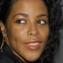} &
		\includegraphics[width=0.087\textwidth]{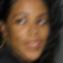} &
		\includegraphics[width=0.087\textwidth]{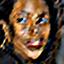} &
		\includegraphics[width=0.087\textwidth]{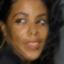} &
		\includegraphics[width=0.087\textwidth]{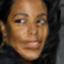} &
		\includegraphics[width=0.087\textwidth]{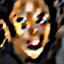} &
		\includegraphics[width=0.087\textwidth]{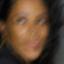} &
		\includegraphics[width=0.087\textwidth]{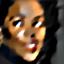} &
		\includegraphics[width=0.087\textwidth]{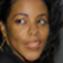} &
		\includegraphics[width=0.087\textwidth]{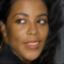} &
		\includegraphics[width=0.087\textwidth]{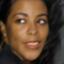} 
		\\
		\includegraphics[width=0.087\textwidth]{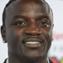} &
		\includegraphics[width=0.087\textwidth]{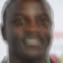} &
		\includegraphics[width=0.087\textwidth]{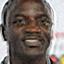} &
		\includegraphics[width=0.087\textwidth]{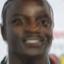} &
		\includegraphics[width=0.087\textwidth]{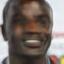} &
		\includegraphics[width=0.087\textwidth]{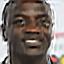} &
		\includegraphics[width=0.087\textwidth]{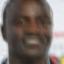} &
		\includegraphics[width=0.087\textwidth]{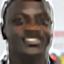} &
		\includegraphics[width=0.087\textwidth]{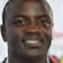} &
		\includegraphics[width=0.087\textwidth]{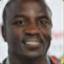} &
		\includegraphics[width=0.087\textwidth]{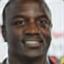} 
		\\
		(a) GT &
		(b) Blurred &
		(c) \citeauthor{DBLP:conf/eccv/PanHSY14} &
		(d) \citeauthor{Nah_2017_CVPR} &
		(e) \citeauthor{tao2018scale} &
		(f) \citeauthor{li2018learning} &
		(g) \citeauthor{deblurgan} &
		(h) \citeauthor{PAN17} &
		(i) \citeauthor{Jin18}&
		(j) \citeauthor{Ziyi-CVPR-2018} &
		(k) Ours 
		\\
		& 
		& 
		\citeyear{DBLP:conf/eccv/PanHSY14} &
		\citeyear{Nah_2017_CVPR} &
		\citeyear{tao2018scale} &
		\citeyear{li2018learning} &
		\citeyear{deblurgan} &
		\citeyear{PAN17} &
		\citeyear{Jin18} &
		\citeyear{Ziyi-CVPR-2018} &
		\\
	\end{tabular}
	\vspace{-1mm}
	\caption{
		\textbf{Visual comparison on images with different attributes.}
		We show that the proposed method is able to generate sharp images and robust to several scenarios, e.g., occlusion with sunglass or hands (row 1 to 3), faces with mustaches (row 1 and 4), over-exposed images (row 5 to 6), and people with different skin colors.
	}
	\label{fig:specific}
\end{figure*}

\section{Evaluation against with the State-of-the-art Methods}
In this section, we present evaluations against the state-of-the-art deblurring approaches in terms of the restoration quality, face detection, face recognition, and execution time.
We also provide visual comparisons on synthetic datasets and real blurred images.
Finally, we discuss the limitation and failure cases of the proposed method.

\subsection{Restoration quality}
We compare the proposed method with the state-of-the-art deblurring algorithms, including MAP-based methods~\citep{DBLP:journals/tog/ChoL09,DBLP:conf/cvpr/KrishnanTF11,DBLP:journals/tog/ShanJA08,DBLP:conf/cvpr/XuZJ13,DBLP:conf/cvpr/ZhongCMPW13,DBLP:conf/eccv/PanHSY14,PAN17,li2018learning} and CNN-based methods~\citep{Nah_2017_CVPR,tao2018scale,deblurgan,Jin18,Ziyi-CVPR-2018}.
We evaluate all the algorithms on both the Helen and CelebA test sets.
\tabref{evaluate_by_kernel} presents the average PSNR and SSIM for different sizes of blur kernels,
and~\tabref{average_psnr_ssim} shows the average and the worst PSNR/SSIM on the entire datasets for each method.
We note that the optimization-based methods~\citep{DBLP:journals/tog/ShanJA08, DBLP:journals/tog/ChoL09, DBLP:conf/cvpr/KrishnanTF11, DBLP:conf/cvpr/XuZJ13, DBLP:conf/cvpr/ZhongCMPW13, DBLP:conf/eccv/PanHSY14, PAN17} may generate severe visual artifacts when the blur kernel is not estimated well and achieve significant lower PSNR/SSIM values.
The proposed method performs favorably against existing deblurring approaches and our preliminary method~\citep{Ziyi-CVPR-2018} on both datasets.

We show the results of the Helen dataset in~\figref{visual_comparison_helen} and the CelebA dataset in~\figref{visual_comparison_celeba}.
Conventional MAP-based approaches~\citep{DBLP:journals/tog/ChoL09,DBLP:conf/cvpr/KrishnanTF11,DBLP:journals/tog/ShanJA08,DBLP:conf/cvpr/XuZJ13,DBLP:conf/cvpr/ZhongCMPW13,PAN17} 
do not estimate blur kernels well and therefore generate more ringing artifacts.
The face deblurring approach~\citep{DBLP:conf/eccv/PanHSY14} is not robust to noise and 
the performance depends heavily on the similarity of the reference image.
There are several ringing artifacts in the deblurred images by~\citet{DBLP:conf/eccv/PanHSY14}.
The method of~\citet{li2018learning} generates sharp debluured images, but the faces do not look realistic.
The CNN-based methods~\citep{Nah_2017_CVPR,deblurgan,tao2018scale,Jin18} do not consider the face semantic information and thus cannot effectively reduce the motion blur.

Both the method by~\citet{Ziyi-CVPR-2018} and the proposed model obtain visually pleasing results.
However, the method by \citet{Ziyi-CVPR-2018} is not robust to the error on semantic labels (which is predicted from blurred images) and less effective in restoring facial details (e.g., the mouth of the first and second rows in~\figref{visual_comparison_celeba}).
In contrast, the proposed method extracts more accurate semantic priors and restores better facial structures and details (e.g., the eyes of the second and third rows in~\figref{visual_comparison_helen}).

In~\figref{specific}, we show the deblurring results from images with specific attributes, such as occlusion, mustaches, saturation, and people with different skin colors.
As our test set does not contain images with significant saturation, we adjust the intensity of the blurred images (row 5 and 6 of \figref{specific}) by multiplying the Y-channel by $1.5 \times$.
The proposed method can still recover more facial details than existing approaches from such an input.
Overall, our method performs well in real-world scenarios.

\begin{figure}
	\centering
	\footnotesize
	\renewcommand{\tabcolsep}{2pt} % adjust horizontal space
        \includegraphics[width=1.0\columnwidth]{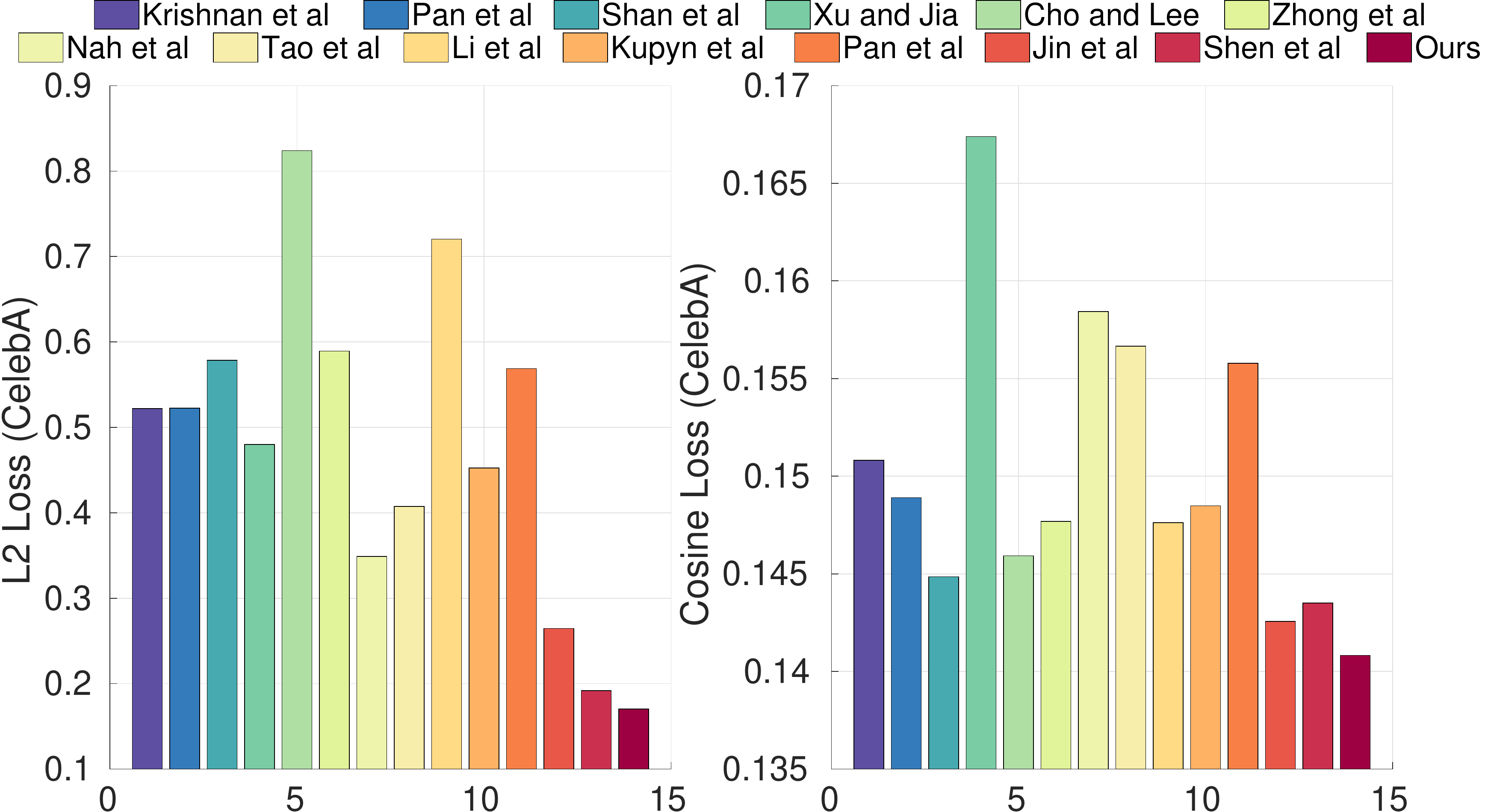}
	%\vspace{1mm}
	\caption{
		\textbf{Quantitative evaluation on face identity.}
		We compute the L2 and cosine losses on the features extracted from the FaceNet~\citep{FaceNet}.
		The proposed method has the lowest values on the CelebA test sets.
	}
	\label{fig:identity_distance}
\end{figure}

\begin{figure*}
	\scriptsize
	\renewcommand{\tabcolsep}{1pt} % adjust horizontal space
	\renewcommand{\arraystretch}{0.8} % adjust vertical space
	\centering
	\begin{tabular}{ccccccccccc}
		\includegraphics[width=0.087\textwidth]{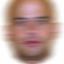} &
		\includegraphics[width=0.087\textwidth]{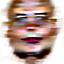} &
		\includegraphics[width=0.087\textwidth]{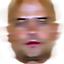} &
		\includegraphics[width=0.087\textwidth]{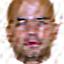} &
		\includegraphics[width=0.087\textwidth]{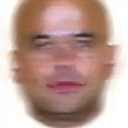} &
		\includegraphics[width=0.087\textwidth]{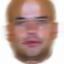} &
		\includegraphics[width=0.087\textwidth]{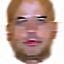} &
		\includegraphics[width=0.087\textwidth]{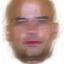} &
		\includegraphics[width=0.087\textwidth]{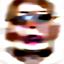} &
		\includegraphics[width=0.087\textwidth]{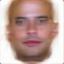} &
		\includegraphics[width=0.087\textwidth]{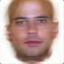} 
		\\
		\includegraphics[width=0.087\textwidth]{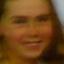} &
		\includegraphics[width=0.087\textwidth]{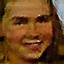} &
		\includegraphics[width=0.087\textwidth]{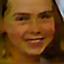} &
		\includegraphics[width=0.087\textwidth]{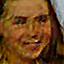} &
		\includegraphics[width=0.087\textwidth]{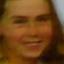} &
		\includegraphics[width=0.087\textwidth]{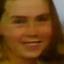} &
		\includegraphics[width=0.087\textwidth]{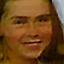} &
		\includegraphics[width=0.087\textwidth]{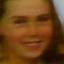} &
		\includegraphics[width=0.087\textwidth]{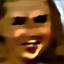} &
		\includegraphics[width=0.087\textwidth]{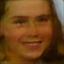} &
		\includegraphics[width=0.087\textwidth]{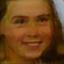} 
		\\
		\includegraphics[width=0.087\textwidth]{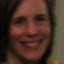} &
		\includegraphics[width=0.087\textwidth]{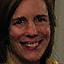} &
		\includegraphics[width=0.087\textwidth]{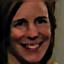} &
		\includegraphics[width=0.087\textwidth]{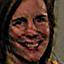} &
		\includegraphics[width=0.087\textwidth]{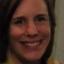} &
		\includegraphics[width=0.087\textwidth]{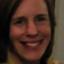} &
		\includegraphics[width=0.087\textwidth]{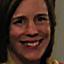} &
		\includegraphics[width=0.087\textwidth]{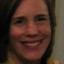} &
		\includegraphics[width=0.087\textwidth]{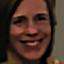} &
		\includegraphics[width=0.087\textwidth]{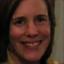} &
		\includegraphics[width=0.087\textwidth]{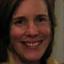} 
		\\
		\includegraphics[width=0.087\textwidth]{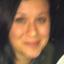} &
		\includegraphics[width=0.087\textwidth]{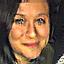} &
		\includegraphics[width=0.087\textwidth]{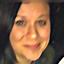} &
		\includegraphics[width=0.087\textwidth]{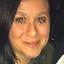} &
		\includegraphics[width=0.087\textwidth]{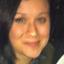} &
		\includegraphics[width=0.087\textwidth]{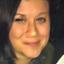} &
		\includegraphics[width=0.087\textwidth]{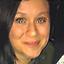} &
		\includegraphics[width=0.087\textwidth]{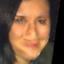} &
		\includegraphics[width=0.087\textwidth]{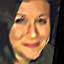} &
		\includegraphics[width=0.087\textwidth]{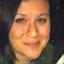} &
		\includegraphics[width=0.087\textwidth]{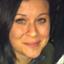} 
		\\
		\includegraphics[width=0.087\textwidth]{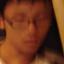} &
		\includegraphics[width=0.087\textwidth]{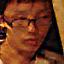} &
		\includegraphics[width=0.087\textwidth]{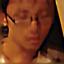} &
		\includegraphics[width=0.087\textwidth]{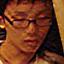} &
		\includegraphics[width=0.087\textwidth]{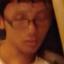} &
		\includegraphics[width=0.087\textwidth]{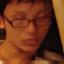} &
		\includegraphics[width=0.087\textwidth]{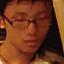} &
		\includegraphics[width=0.087\textwidth]{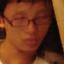} &
		\includegraphics[width=0.087\textwidth]{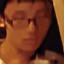} &
		\includegraphics[width=0.087\textwidth]{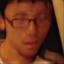} &
		\includegraphics[width=0.087\textwidth]{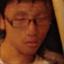} 
		\\
		\includegraphics[width=0.087\textwidth]{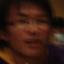} &
		\includegraphics[width=0.087\textwidth]{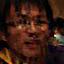} &
		\includegraphics[width=0.087\textwidth]{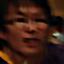} &
		\includegraphics[width=0.087\textwidth]{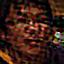} &
		\includegraphics[width=0.087\textwidth]{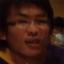} &
		\includegraphics[width=0.087\textwidth]{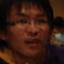} &
		\includegraphics[width=0.087\textwidth]{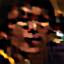} &
		\includegraphics[width=0.087\textwidth]{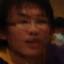} &
		\includegraphics[width=0.087\textwidth]{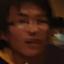} &
		\includegraphics[width=0.087\textwidth]{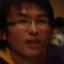} &
		\includegraphics[width=0.087\textwidth]{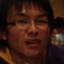} 
		\\
		(a) Blurred &
		(b) \citeauthor{DBLP:conf/cvpr/XuZJ13} &
		(c) \citeauthor{DBLP:conf/cvpr/ZhongCMPW13} &
		(d) \citeauthor{DBLP:conf/eccv/PanHSY14} &
		(e) \citeauthor{Nah_2017_CVPR} &
		(f) \citeauthor{tao2018scale} &
		(g) \citeauthor{li2018learning} &
		(h) \citeauthor{deblurgan} &
		(i) \citeauthor{PAN17} &
		(i) \citeauthor{Ziyi-CVPR-2018} &
		(j) Ours 
		\\
		& 
		\citeyear{DBLP:conf/cvpr/XuZJ13} &
		\citeyear{DBLP:conf/cvpr/ZhongCMPW13} &
		\citeyear{DBLP:conf/eccv/PanHSY14} &
		\citeyear{Nah_2017_CVPR} &
		\citeyear{tao2018scale} &
		\citeyear{li2018learning} &
		\citeyear{deblurgan} &
		\citeyear{PAN17} &
		\citeyear{Ziyi-CVPR-2018} &
		\\
	\end{tabular}
	\vspace{-1mm}
	\caption{
		\textbf{Visual comparison on real blurred images.}
		The proposed method generates visually pleasing deblurred results with fewer artifacts.
	}
	\label{fig:comparison_real}
\end{figure*}

\subsection{Face recognition}
We also demonstrate the performance of the proposed method by evaluating the face identity distance, face detection, and recognition accuracy.

\Paragraph{Identity distance.}
We use the FaceNet~\citep{FaceNet} to extract face features and compute the identity distance with the $L_2$ loss and cosine loss~\citep{wang2018cosface} between the ground truth image and deblurred image.
\figref{identity_distance} shows that the deblurred images from the proposed method have the lowest identity distance on both measurements, which demonstrates that the proposed method preserves the face identity well.

\Paragraph{Face detection.}
We use the OpenFace toolbox~\citep{openface} to detect the face for each image in the CelebA test set.
We show the success rate of the face detection for blurred images and the state-of-the-art deblurring approaches in~\tabref{celebA}.
The clear images have a success rate of $96\%$, while the success rate on blurred images drops 
to $77.4\%$ due to motion blur.
The deblurred images from some of the evaluated methods have an 
lower success rate as the images contain severe ringing artifacts.
In contrast, the proposed method has $95.3\%$ success rate, which is close to the upper bound of the clear images.

\Paragraph{Face recognition.}
As the CelebA dataset contains identity labels, we conduct another experiment on the identity recognition.
We consider our CelebA test images as a probe set, which has 100 different identities.
For each identity, we collect additional 9 clear face images as a gallery set.
For each image in the probe set, our goal is to find the most similar face image from the gallery set and identify whether they belong to the same identity.

Given a blurred or deblurred image from the probe set, we compute the identity distance with all images in the gallery set and select the top-$K$ nearest matches.
\tabref{celebA} shows the top-1, top-3 and top-5 accuracy.
The proposed method generates fewer artifacts and thus achieves the highest recognition accuracy against other evaluated approaches.

\begin{table}
	\centering
	\footnotesize
	\caption{
		\textbf{Face detection and recognition on the CelebA dataset.}
		We show the success rate of face detection and top-1, top-3 and top-5 accuracy of face recognition.
	}
	\vspace{-1mm}
	\label{tab:celebA}
	\begin{center}
		\begin{tabular}{r|c|c|c|c}
			\toprule
			Method &
			Detection &
			Top-1 &
			Top-3 &
			Top-5
			\\
			\midrule
			Clear images &
			96.0$\%$ & 74.0$\%$ & 86.4$\%$ & 90.0$\%$
			\\
			Blurred images &
			77.4$\%$ & 29.1$\%$ & 43.4$\%$ & 51.3$\%$
			\\
			\midrule
			\citet{DBLP:journals/tog/ShanJA08} &
			76.0$\%$ & 32.4$\%$ & 46.9$\%$ & 54.0$\%$
			\\
			\citet{DBLP:journals/tog/ChoL09} &
			52.2$\%$ & 17.2$\%$ & 27.3$\%$ & 32.5$\%$
			\\
			\citet{DBLP:conf/cvpr/KrishnanTF11} &
			80.0$\%$ & 33.8$\%$ & 48.9$\%$ & 56.6$\%$
			\\
			\citet{DBLP:conf/cvpr/XuZJ13} &
			82.5$\%$ & 41.1$\%$ & 55.4$\%$ & 62.1$\%$
			\\
			\citet{DBLP:conf/cvpr/ZhongCMPW13} &
			69.5$\%$ & 27.6$\%$ & 41.6$\%$ & 48.5$\%$
			\\
			\citet{DBLP:conf/eccv/PanHSY14} &
			78.9$\%$ & 42.0$\%$ & 55.7$\%$ & 62.2$\%$
			\\
			\citet{PAN17} &
			74.3$\%$ & 40.9$\%$ & 48.2$\%$ & 58.3$\%$
			\\		
			\citet{Nah_2017_CVPR} &
			86.0$\%$ & 40.1$\%$ & 55.3$\%$ & 62.4$\%$
			\\
			\citet{tao2018scale} &
			80.5$\%$ & 36.8$\%$ & 53.6$\%$ & 59.8$\%$
			\\
			\citet{li2018learning} &
			78.2$\%$ & 40.7$\%$ & 49.8$\%$ & 56.2$\%$
			\\		
			\citet{deblurgan} &
			88.4$\%$ & 43.5$\%$ & 59.6$\%$ & 65.3$\%$
			\\
			\citet{Jin18} &
			89.8$\%$&42.3$\%$&60.2$\%$&67.7$\%$\\
			\citet{Ziyi-CVPR-2018} &
			\second{94.8$\%$} & \second{48.3$\%$} & \second{63.2$\%$} & \second{70.0$\%$}	
			\\
			Ours &
			\first{95.3$\%$} & \first{53.8$\%$} & \first{68.7$\%$} & \first{74.2$\%$}
			\\
			\bottomrule
		\end{tabular}
	\end{center}
	%\vspace{-3mm}
\end{table}

\subsection{Real-world blurred images}
We evaluate the proposed method on face images collected from the real blurred dataset of~\citet{Lai-CVPR-2016}.
As real images usually contain outliers that cannot be modeled well by Gaussian distributions, conventional methods fail to estimate the blur kernel and generate serious ringing artifacts.
The CNN-based generic deblurring method~\citep{Nah_2017_CVPR} generates overly smooth results.
In contrast, both the method of~\citet{Ziyi-CVPR-2018} and the proposed model restore sharp and visually pleasing face images.

\subsection{Execution time}
We evaluate the execution time of the state-of-the-art approaches and the proposed model on a machine with a 3.4 GHz Intel i7 CPU (64G RAM) and an NVIDIA Titan X GPU card (12G memory).
\tabref{runtime} shows the average execution time based on 10 images with a size of $128 \times 128$.
Most conventional approaches require solving several iterative optimization problems and therefore are computationally expensive.
Since we use only two scales and fewer residual blocks, our model is more efficient than the model of~\citet{Nah_2017_CVPR}.
The proposed model is slightly slower than the model of~\citet{Ziyi-CVPR-2018} as there is an additional coarse deblurring network.

\begin{table}
	\centering
	\footnotesize
	\caption{\textbf{Comparison of execution time and model size.}
		We report the average execution time on 10 images with the size of $128 \times 128$. }
	\label{tab:runtime}
	\vspace{1mm}
	\resizebox{1.0\columnwidth}{!}{
	\begin{tabular}{r|c|c|c}
		\toprule
		Method & Implementation & Seconds & Parameters \\
		\midrule
		\citet{DBLP:journals/tog/ShanJA08}
		& C++ (CPU) & 16.32 & $-$\\
		\citet{DBLP:journals/tog/ChoL09}
		& C++ (CPU) & 0.41 & $-$\\
		\citet{DBLP:conf/cvpr/KrishnanTF11}
		& MATLAB (CPU) & 2.52 & $-$\\
		\citet{DBLP:conf/cvpr/XuZJ13}
		& C++ (CPU) & 0.31 & $-$\\
		\citet{DBLP:conf/cvpr/ZhongCMPW13}
		& MATLAB (CPU) & 8.07 & $-$\\
		\citet{DBLP:conf/eccv/PanHSY14}
		& MATLAB (CPU) & 8.11 & $-$\\
		\citet{PAN17}
		& MATLAB (CPU) & 10.55 & $-$\\
		\citet{Nah_2017_CVPR}
		& MATLAB (GPU) & 0.09 & 303.6M \\
		\citet{tao2018scale}
		& Python (GPU) & 0.15 &32.2M\\
		\citet{li2018learning}
		& MATLAB (CPU) & 18.53& 558K \\
		\citet{deblurgan}
		& Python (GPU) & 0.05 &45.5M\\
		\citet{Jin18}
		& Torch (GPU) & 0.01&1.4M\\
		\citet{Ziyi-CVPR-2018}
		& MATLAB (GPU) & 0.05&14.8M \\
		Ours
		& MATLAB (GPU) & 0.08 &26.6M\\
		\bottomrule
	\end{tabular}
}
\end{table}

\subsection{Limitations and discussions}

\begin{figure}
	\scriptsize
	\renewcommand{\tabcolsep}{1pt} % adjust horizontal space
	\renewcommand{\arraystretch}{1.0} % adjust vertical space
	\centering
	\begin{tabular}{ccccc}
	    \includegraphics[width=0.19\columnwidth]{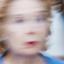} &
        \includegraphics[width=0.19\columnwidth]{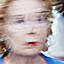} &
        \includegraphics[width=0.19\columnwidth]{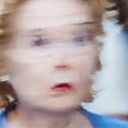} &
        \includegraphics[width=0.19\columnwidth]{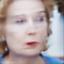} &
        \includegraphics[width=0.19\columnwidth]{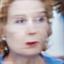}
        \\
        \includegraphics[width=0.19\columnwidth]{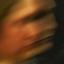} &
        \includegraphics[width=0.19\columnwidth]{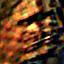} &
        \includegraphics[width=0.19\columnwidth]{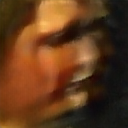} &
        \includegraphics[width=0.19\columnwidth]{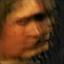} &
        \includegraphics[width=0.19\columnwidth]{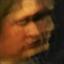}
        \\
		(a) Blurred &
		(b) \citeauthor{DBLP:conf/eccv/PanHSY14} &
		(c) \citeauthor{Nah_2017_CVPR} &
		(d) \citeauthor{Ziyi-CVPR-2018}&
		(e) Ours
		\\
		images &
		\citeyear{DBLP:conf/eccv/PanHSY14} &
		\citeyear{Nah_2017_CVPR} &
		\citeyear{Ziyi-CVPR-2018} &
	\end{tabular}
	\vspace{-1mm}
	\caption{
		\textbf{Failure cases.} 
		Our method fails when the input image suffers from extremely large motion blur and the semantic labels cannot be estimated well.
    \label{fig:fail}
	}
\end{figure}

\begin{figure}
	\scriptsize
	\renewcommand{\tabcolsep}{1pt} % adjust horizontal space
	\renewcommand{\arraystretch}{0.8} % adjust vertical space
	\centering
	\begin{tabular}{ccccccc}
		\includegraphics[width=0.134\columnwidth]{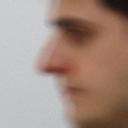} &
		\includegraphics[width=0.134\columnwidth]{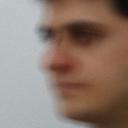} &
		\includegraphics[width=0.134\columnwidth]{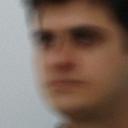} &
		\includegraphics[width=0.134\columnwidth]{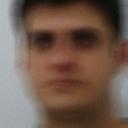} &
		\includegraphics[width=0.134\columnwidth]{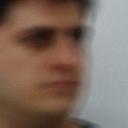} &
		\includegraphics[width=0.134\columnwidth]{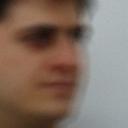} &
		\includegraphics[width=0.134\columnwidth]{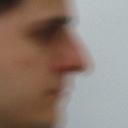} 
		\\
		\multicolumn{7}{c}{(a) Blurred images}
		\\
		\includegraphics[width=0.134\columnwidth]{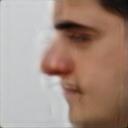} &
		\includegraphics[width=0.134\columnwidth]{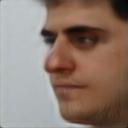} &
		\includegraphics[width=0.134\columnwidth]{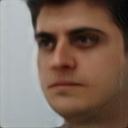} & 
		\includegraphics[width=0.134\columnwidth]{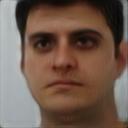} &
		\includegraphics[width=0.134\columnwidth]{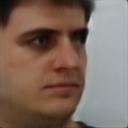} &
		\includegraphics[width=0.134\columnwidth]{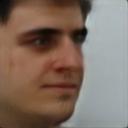} &
		\includegraphics[width=0.134\columnwidth]{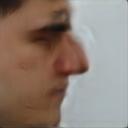} 
		\\
		\multicolumn{7}{c}{(b) Our deblurred results}
		\\
		\includegraphics[width=0.134\columnwidth]{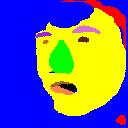} &
		\includegraphics[width=0.134\columnwidth]{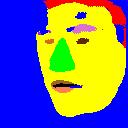} &
		\includegraphics[width=0.134\columnwidth]{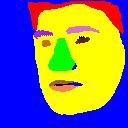} & 
		\includegraphics[width=0.134\columnwidth]{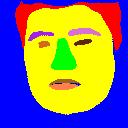} &
		\includegraphics[width=0.134\columnwidth]{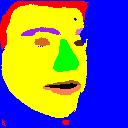} &
		\includegraphics[width=0.134\columnwidth]{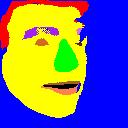} &
		\includegraphics[width=0.134\columnwidth]{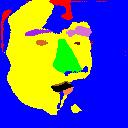} 
		\\
		\multicolumn{7}{c}{(c) Semantic labels}
	\end{tabular}
	\vspace{-1mm}
	\caption{
		\textbf{Deblurring profile faces.} 
		We evaluate our model on the FEI face database~\citep{thomaz2010new}.
		The proposed model becomes less effective when a face is rotated by 90 degrees.
	}
	\label{fig:profile}
\end{figure}

Our method is likely to fail in two situations.
First, when the input image contains severe non-uniform blur or non-Gaussian noise, our model may not be able to reduce the blur effectively, as shown in~\figref{fail}.
A potential solution is to synthesize more training data with complex motion models or realistic noise~\citep{foi2008practical}.
Second, when the face cannot be well aligned (e.g., profile faces in~\figref{fail} bottom), the face parsing network may not estimate accurate semantic labels to guide the deblurring network.
To further analyze the performance of the proposed model on profile faces, we evaluate the face images from the FEI face database~\citep{thomaz2010new}, where each face is captured under different rotation angles.
As shown in~\figref{profile}, our model performs well on frontal faces and profile faces which are rotated by about $60$ degrees (i.e., $2^{\text{nd}}$ to $6^{\text{th}}$ columns of~\figref{profile}).
For extreme cases (e.g., rotated by about 90 degrees as shown in the $1^{\text{st}}$ and $7^{\text{th}}$ columns of~\figref{profile}), our deblurred results contain some visual artifacts around the nose and mouth.
The eyes are not restored well due to the inaccurate semantic labels.

\section{Conclusions}
\label{sec:conclusions}

In this work, we propose a multi-scale deep convolutional neural network for face image deblurring.
We exploit the face semantic information as global priors and local structural constraints to better restore the shape and detail of face images.
Compared with the preliminary work~\citep{Ziyi-CVPR-2018} which obtains the semantic labels from the input blurred image, we show that the semantic information extracted from a coarse deblurred image is more accurate and leads to better performance on deblurring images.
Furthermore, we propose an adaptive local structural loss to balance the weights of facial key components and restore better content and details.
Experimental results on image deblurring, execution time and face recognition demonstrate that the proposed method performs favorably against our preliminary method~\citep{Ziyi-CVPR-2018} and the state-of-the-art deblurring algorithms.

\begin{acknowledgements}
This work was supported by the Major Science Instrument Program of the National Natural Science Foundation of China under Grant 61527802, the General Program of National Nature Science Foundation of China under Grants 61371132 and 61471043, NSF CAREER (No. 1149783) and gifts from Adobe and Nvidia.
\end{acknowledgements}

%\bibliographystyle{deblur}      % basic style, author-year citations

%\clearpage

%\bibliographystyle{spmpsci}
%	\bibliography{deblur}
{\small
\bibliographystyle{spbasic}
\bibliography{deblur}
}
\end{document}